\documentclass[12pt]{iopart}
\usepackage{comment,mfirstuc}
\usepackage{graphicx}% Include figure files
\usepackage{grffile}
\usepackage{algorithm}% http://ctan.org/pkg/algorithm
\usepackage{algpseudocode}% http://ctan.org/pkg/algorithmicx
\usepackage{hyperref}% add hypertext capabilities
\usepackage{soul}
\usepackage{listings}
\usepackage{multirow}
\usepackage{dcolumn}% Align table columns on decimal point
\usepackage{bm}% bold math
\usepackage{placeins}
\expandafter\let\csname equation*\endcsname\relax
\expandafter\let\csname endequation*\endcsname\relax
\usepackage{amsmath}
\usepackage{amssymb}
\usepackage{mathtools}
\usepackage{subfig}
\usepackage[bottom=1.0in]{geometry}
\usepackage{cite}
\usepackage{lineno}
\RequirePackage{xcolor}
\usepackage[misc]{ifsym}
\usepackage{tcolorbox} % Create coloured boxes (e.g. the one for the key-words)
\usepackage{stfloats} % Correct position of the tables
\usepackage{siunitx}
\usepackage{colortbl}
\usepackage{booktabs} % Required for \toprule, \midrule, and \bottomrule

\usepackage{array}
\newcolumntype{P}[1]{>{\centering\arraybackslash}p{#1}}
\newcolumntype{M}[1]{>{\centering\arraybackslash}m{#1}}
\hypersetup{ 
    pdfnewwindow=true,      % links in new window
    colorlinks=true,       % false: boxed links; true: colored links
    linkcolor=blue,         % color of internal links
    citecolor=blue,        % color of links to bibliography
    filecolor=blue,      % color of file links
    urlcolor=blue        % color of external links
}  

\algblock{Input}{EndInput}
\algnotext{EndInput}
\algblock{Output}{EndOutput}
\algnotext{EndOutput}

\def\be{\begin{eqnarray} &&} 
 
\def\ee{\end{eqnarray}}

\makeatletter
\newcommand{\mainmatter}{%
  \setcounter{footnote}{0}%
  \patchcmd{\@makefntext}{\fnsymbol}{\arabic}{}{}%
  \patchcmd{\@thefnmark}{\fnsymbol}{\arabic}{}{}%
  \def\@makefnmark{\textsuperscript{\arabic{footnote}}}
  \long\def\@makefntext##1{\parindent 1em\noindent
        \hb@xt@1.8em{%
            \hss\@textsuperscript{\normalfont\@thefnmark}}##1}%
}
\makeatother

\newcommand{\addComment}[2]{
  \expandafter\newcommand\csname #1\endcsname[1]{{\bf \color{#2} \capitalisewords{#1}:\,##1}}
  \expandafter\newcommand\csname #1cor\endcsname[2]{{\color{#2} \capitalisewords{#1}:\,\st{##1}{\bf ##2}}}
  \expandafter\newcommand\csname #1color\endcsname{#2}
}
\addComment{cris}{blue} 
\addComment{james}{red}
\addComment{zisis}{brown}

\begin{document}

% Keywords command
\providecommand{\keywords}[1]
{
  \small
  \textbf{Keywords:}  {\color{blue}#1
  }
}

%\title[\scriptsize{Harnessing Deep Learning for Event-Level Uncertainty Quantification in Deep Inelastic Scattering at the EIC}]{Harnessing Deep Learning for Event-Level Uncertainty Quantification in Deep Inelastic Scattering at the EIC}  

%\title[\scriptsize{Event-Level Uncertainty Quantification using Physics-Informed Bayesian Neural Networks with Flow approximated Posteriors}]{ELUQuant: harnessing Event-Level Uncertainty Quantification in Deep Inelastic Scattering}

\title[\scriptsize{ELUQuant: Event-Level Uncertainty Quantification in Deep Inelastic Scattering}]{ELUQuant: Event-Level Uncertainty Quantification in Deep Inelastic Scattering}

%\title[\scriptsize{}]{ELUQuant: Event-Level Uncertainty Quantification
%using Physics-Informed Bayesian Neural Networks
%with Flow approximated Posteriors --- A Deep Inelastic Scattering Study}

%Diversity Identification
%\cris{One-Class Classification}}

%A Conditional Generative Approach to Anomaly Detection
%{Flux \& Mutability: A Conditional Generative Approach to Anomaly Detection \cris

\author{C. Fanelli$^{{\ast}}$, J. Giroux$^{{\ast}}$} %\S, %x$^{5,\star}$, Z. Papandreou$^{5,\ddagger}$}

\address{
Department of Data Science, William \& Mary, Williamsburg, VA 23185, USA\\
}

\ead{{\color{blue}
$^{\ast}$\{cfanelli, jgiroux\}@wm.edu,
%$^{\S}$email@wm.edu,
%$^{\star}$ 
%$^{\ddagger}$
}}

\vspace{10pt}
\begin{indented}
\item[]\today
\end{indented}

%\linenumbers

\begin{abstract}

We introduce a physics-informed Bayesian Neural Network (BNN)
with flow approximated posteriors using multiplicative normalizing flows (MNF) for detailed uncertainty quantification (UQ) at the physics event-level. Our method is capable of identifying both heteroskedastic aleatoric and epistemic uncertainties, providing granular physical insights. Applied to Deep Inelastic Scattering (DIS) events, our model effectively extracts the kinematic variables $x$, $Q^2$, and $y$, matching the performance of recent deep learning regression techniques but with the critical enhancement of event-level UQ. 
  This detailed description of the underlying uncertainty proves invaluable for decision-making, especially in tasks like event filtering. It also allows for the reduction of true inaccuracies without directly accessing the ground truth. 
  A thorough DIS simulation using the H1 detector at HERA indicates possible applications for the future EIC. Additionally, this paves the way for related tasks such as data quality monitoring and anomaly detection. 
  Remarkably, our approach effectively processes large samples at high rates.

%{\color{blue}Keywords: \textbf{conditional masked autoregressive flow, clustering, anomaly detection, one-class, neutral showers, jets}}

\end{abstract}


%
% Uncomment for keywords
%\vspace{2pc}
%\noindent{\it Keywords}: XXXXXX, YYYYYYYY, ZZZZZZZZZ
%
% Uncomment for Submitted to journal title message
%\submitto{\JPA}
%
% Uncomment if a separate title page is required
%\maketitle
% 
% For two-column output uncomment the next line and choose [10pt] rather than [12pt] in the \documentclass declaration
%\ioptwocol
%

\mainmatter

\section{Introduction}\label{sec:intro}

In experimental nuclear (NP) and high-energy physics (HEP), data analyses typically regress fundamental quantities from observables measured in events produced and detected by experiments. A crucial aspect of these analyses is the corresponding event-level uncertainty quantification (UQ). The method introduced in this work ELUQuant (dubbed ELUQ), to our knowledge, pioneers this in NP/HEP by gleaning insights from computer vision \cite{pmlr-v70-louizos17a} and Multiplicative Normalizing Flows (MNF) in Bayesian Neural Networks (BNN)\cite{kendall2017uncertainties}, 
effectively capturing both heteroskedastic aleatoric and epistemic uncertainties, which influence the regression of fundamental quantities from measured observables.
Deep inelastic scattering (DIS) has recently benefited from deep learning (DL) techniques. An innovative study by Diefenthaler et al.~\cite{diefenthaler2022deeply} employed deep neural networks (DNN) to infer kinematic variables \( Q^2 \) and \( x \) of neutral current DIS from traditional reconstruction methods enhanced through correlations revealed in the simulated datasets of the ZEUS experiment. 
Successively, Arratia \cite{ARRATIA2022166164} applied DNNs, capitalizing on full kinematic information of both the scattered electron and the hadronic-final state, to reconstruct the kinematics of neutral-current DIS events, using H1 experiment simulations. 
Though both papers signify critical advancements in leveraging DNN for DIS, they did not delve into the domain of UQ. Our endeavor with ELUQ distinctively bridges this aspect, and highlights the potential of detailed event-level UQ, a novelty among the referenced works. This methodology harbors promise for any physics analysis demanding nuanced UQ.

This manuscript is structured as follows: Sec. \ref{sec:DIS_UQ} introduces the DIS kinematics, the chosen case study, and both the quantified uncertainty sources (aleatoric and epistemic); Sec. \ref{sec:ELUQuant} delves into the ELUQ architecture, detailing its loss function, training procedures, and inference performance; Sec. \ref{sec:Results} reports the results we obtained using H1's neutral current DIS Monte Carlo dataset also used in \cite{ARRATIA2022166164}.
%\footnote{This data was made available at AI4EIC 2022 in the tutorial provided by Torales Acosta and Mikuni \cite{torales2022omnifold}.} 
Conclusively, Sec. \ref{sec:Impacts} evaluates the broader impacts, emphasizing the effectiveness of ELUQ in event-level UQ and its potential applications for data quality monitoring and anomaly detection.

\section{Kinematic Reconstruction of DIS}\label{sec:DIS_UQ}

\begin{figure}[!]
    \centering
    \includegraphics[width=0.54\textwidth, trim=6cm 5cm 5cm 2cm, clip]{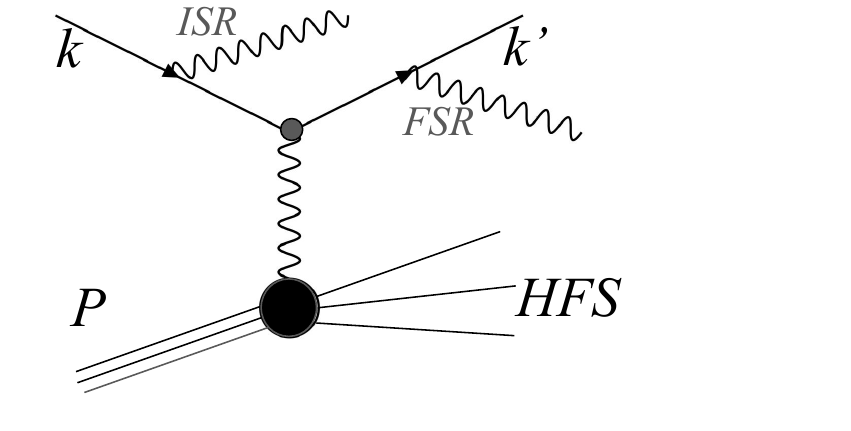} 
    \hspace{-1cm}
    \caption{Neutral current DIS diagram with possible initial and final state radiation. Additional effects may arise from, \textit{e.g.}, higher-order QED corrections at the lepton vertex or QCD corrections (see \cite{diefenthaler2022deeply, ARRATIA2022166164}).}
    \label{fig:DIS_feynman}
\end{figure}

\paragraph{Deep Inelastic Scattering} 
DIS is a reaction used to probe the internal structure of hadrons. In this process, high-energy leptons are scattered off hadrons, revealing intricate details about quarks and gluons~\cite{devenish2011deep}.
Historically, the experiments conducted at the HERA collider, which remains the only electron-proton collider ever constructed, have been instrumental in DIS studies~\cite{ABT1997310, abramowicz2015combination}. The forthcoming Electron Ion Collider (EIC) \cite{khalek2022science} promises to venture into previously uncharted regions of the DIS kinematic spectrum.
Fig. \ref{fig:DIS_feynman} depicts the DIS process, where $k, k'$ and $P$ are the four-momenta of the electrons and proton, respectively, and HFS is the hadronic final state.

DIS kinematics involve: squared four-momentum transfer $Q^2 = -q^{2}=(k-k')^{2}$; inelasticity $y = \frac{q\cdot P}{k\cdot P}$, indicating the electron's energy fraction transferred to the nucleon; and Bjorken scaling $x = \frac{Q^{2}}{sy}$, showing the momentum fraction carried by the struck quark. 
The kinematic variables are related by 
%
%\begin{equation}\label{eq:physics}
 $Q^{2} = sxy$,
%\end{equation}
%
where \( s=(k+P)^{2} \) represents the squared center-of-mass energy.
Momentum and energy conservation in DIS kinematics provide the ability to calculate $x$, $Q^{2}$, and $y$ from measurements. Classical methods for their reconstruction differ (\textit{cf.} \cite{bassler1995kinematic,diefenthaler2022deeply,ARRATIA2022166164}). We compare our results with methods such as Electron (EL), Double Angle (DA), and Jacquet Blondel (JB). 
As highlighted in \cite{diefenthaler2022deeply, ARRATIA2022166164}, the DIS process can be influenced by several factors, such as initial-state and final-state radiation (ISR, FSR). Moreover, higher order QED and QCD corrections can also manifest in the process. 
Each reconstruction method has its strengths across the phase space and sensitivities to radiative effects. For instance, EL uses only measurements of the scattered lepton and excels in high $y$ scenarios but falters at low $y$. In contrast, JB uses only the HFS and performs better at low $y$. 
Hence, regression linking measured quantities to true kinematics is crucial. The true values of $x$, $Q^{2}$, and $y$ in our data are derived from generator-level particle 4-vectors, considering effects like ISR and FSR radiation.

\paragraph{Synthetic dataset and network input}
We utilize full simulation from the H1 experiment that encompass QED radiation and Lund hadronization model.\footnote{The same dataset has been utilized in \cite{ARRATIA2022166164}.}
Table \ref{tab:datasets} summarizes the dataset statistics and size on disk.
\begin{table}[!]
    \centering
    \begin{tabular}{c c c c c}
    \toprule
        Dataset & Training Events & Validation Events & Testing Events & Size on Disk  \\
    \midrule
        H1      & $8.7 \times 10^6$ & $1.9 \times 10^6$ & $1.9 \times 10^6$ & 8 GB\\
        %Athena      & $14 \times 10^6$ & $3 \times 10^6$ & $3 \times 10^6$ & 14 GB \\
    \bottomrule
    \end{tabular}
    \caption{Dataset statistics and size on disk.}
    \label{tab:datasets}
\end{table}
A total of 15 measured input features are used in our work and are sourced from \cite{ARRATIA2022166164}. These encompass seven features sensitive to QED radiation: $p^{bal}_{T} = 1-\frac{p_{T,e}}{T}$ with $T$ as the HFS transverse momentum and $p_{T}$ the electron's; $p^{bal}_{z} = 1 - \frac{\Sigma_{e} + \Sigma}{2E_{0}}$, where $\Sigma_{e}=E-p_{z,e}$ and $\Sigma = \sum^{HFS}_{i} (E_{i}-p_{z,i})$; energy, $\eta$, and $\Delta\phi$ of the nearest photon to the electron beam direction, where $\Delta\phi$ is relative to the electron; $E^{sum}_{CAL}/p_{e}$, the ECAL energy sum within a cone of $\Delta R <$0.4 around the scattered electron; and the count of ECAL clusters within $\Delta R <$0.4.
These seven are merged with another eight: scattered electron's $p_{T,e}, p_{z,e}, E$; the HFS four-vector components $T$, $P_{z,h}$, and $E_{h}$; $\Delta\phi(e,h)$, the angle between the scattered electron and HFS momentum; and the difference, $\Delta\Sigma=\Sigma_{e}-\Sigma$. 

%\clearpage
\section{ELUQuant Architecture}\label{sec:ELUQuant} 

In this work, ELUQ is applied to the DIS simulated dataset of H1, extracting $x$, $Q^2$, $y$, and their related epistemic and aleatoric uncertainties from 15 measured input features.  Epistemic uncertainty stems from knowledge gaps, improving with more data and refined models. Aleatoric uncertainty, on the other hand, arises from inherent system variability and remains unaffected by additional data. 

\begin{figure}
    \centering
    \includegraphics[width=0.65\textwidth, trim=0cm 0cm 0cm 0cm]{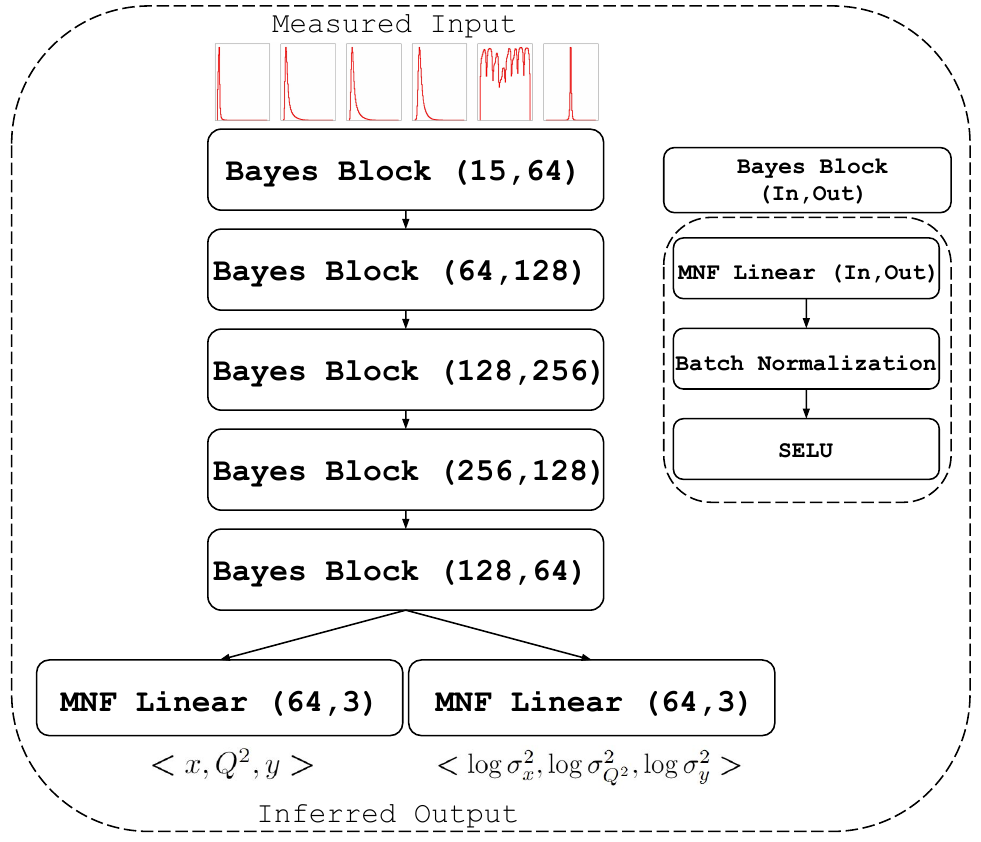}
    \caption{ELUQuant is a bicephalous regression network with Bayesian blocks characterized by MNF modules to approximate posteriors for event-level UQ.
    }
    \label{fig:architecture}
\end{figure}
ELUQ is a bicephalous regression network with Bayesian blocks characterized by MNF to approximate posteriors for event-level UQ. A representation of the architecture is reported in Fig. \ref{fig:architecture}. 
This enables building posteriors over the weights at each layer. Thus, when sampling, the result is a diverse combination of weights within the network. After training, each forward pass through the network yields a distinct set of weights drawn from the learned posterior.

To effectively handle uncertainty in Bayesian networks, the objective is to calculate a posterior, $q(\bold{w}|D)$, where $\bold{w}$ are the weights of the NN. This allows predictions on the regressed quantities through a posterior distribution $q(\bold{y}|\bold{x},D)$, integrated over the space of the weights. However, the formulation of such a posterior is intractable and therefore Bayesian inference must be employed.
Typically, fully factorized Gaussians are assumed as an approximate posterior $q(\bold{W})$ such that we can minimize the evidence lower bound between the approximated posterior and the assumed prior. This is generally limiting and can underestimate true uncertainty. Another method is to utilize random auxiliary variables to improve the approximate posterior via a mixing density. In Louizos et al. \cite{pmlr-v70-louizos17a}, they parameterize the mixing density in terms of auxiliary variables $\bold{z}$, which are in turn parameterized by normalizing flows to allow flexibility and local reparameterizations. They reduce the computational overhead of using a normalizing flow by allowing \(\mathbf{z}\) to act multiplicatively on the means.

Given a set of Gaussian weights, the pre-activation of neurons can be considered as a linear combination of the weights, which is in itself Gaussian. Louizos et al. \cite{pmlr-v70-louizos17a} further show that doing this for each injection within a mini-batch results in a different set of weights, lower variance in gradients, and an overall more stable optimization. An algorithmic description can be found in \cite{pmlr-v70-louizos17a} (Algorithm 1).

\paragraph{Loss function}
The total loss function is the sum of different contributions: 
\begin{equation}
    \mathcal{L}_{Tot.} = \mathcal{L}_{Reg.} + \alpha \mathcal{L}_{Phys.} + \beta \mathcal{L}_{MNF.}
\end{equation}
The regression loss, Eq. (\ref{eq:regression_loss}), provides the DIS kinematic vector of observables we want to predict, namely $\bold{v}=$($x$,$Q^2$,$y$), as well as the corresponding heteroskedastic aleatoric term $\boldsymbol{\sigma}=$($\sigma(x)$,$\sigma(Q^{2})$,$\sigma(y)$):
\begin{equation}\label{eq:regression_loss}
    \mathcal{L}_{Reg.} = \frac{1}{N}\sum_i\sum_j \frac{1}{2}( e^{-\bold{s_j}} \|\bold{v}_j - \bold{\hat{v}}_j\|^2 + \bold{s}_j), \; \bold{s}_j = \log \boldsymbol{\sigma}_j^2
\end{equation}
The sum $i$ runs over all vectors in the mini-batch, and the sum $j$ runs over elements in the vector, where the epistemic term is captured by $\|\bold{v}_i - \bold{\hat{v}}_i\|$. The use of a logarithm at network output has been demonstrated in \cite{kendall2017uncertainties} to be more numerically stable than regressing the variance, $\sigma^2$.\footnote{Note that we do not directly apply a logarithm activation, but rather treat linear activations in terms of logarithms.}
Looking closely at Eq. (\eqref{eq:regression_loss}), this is the logarithm of a multivariate normal distribution, see \cite{kendall2017uncertainties}.
%
%
%The use of logarithm has been demonstrated in \cite{kendall2017uncertainties} to be more numerically stable than regressing the variance, $\sigma^2$, as the loss avoids a potential division by zero. They also show the exponential mapping allows regression of unconstrained scalar values, where $e^{-s_i}$ is resolved to the positive domain giving valid values for variance. 
%
The physics-informed term, Eq. (\ref{eq:physics_loss}), is applied on the regressed observables which ideally should match the truth where $Q^2=sxy$ holds:
\begin{equation}\label{eq:physics_loss}
    \mathcal{L}_{Phys.} = \frac{1}{N}\sum_i \log\hat{Q}_i^2 - ( \log s_i + \log\hat{x}_i + \log\hat{y}_i), 
\end{equation}
where the Mandelstam $s$ is calculated at the ground truth level. 
%
%\cris{The Kullback Leibler term encodes the MNF --- matching of distributions --- write better \cite{pmlr-v70-louizos17a}}
%
The Kullback Leibler, Eq. (\ref{eq:lowerbound_loss}), term is adapted from \cite{pmlr-v70-louizos17a}, which employs MNF in variational Bayesian Neural Networks to improve the posterior approximation. 
%We deploy Gaussian priors assuming normally distributed uncertainties. 
%
We deploy Gaussian priors, and the posterior distribution is given by the product of a Gaussian and mixing density parameterized by a normalizing flow. As shown in \cite{pmlr-v70-louizos17a}, this parameterization is flexible, allowing nonlinear and multimodal dependencies between the weight elements.
%
%\cris{encodes the similarity of the learned weight posterior to the weight prior. (?)}
%
%\begin{equation}\label{eq:lowerbound_loss}
%    \mathcal{L}_{NF} = -KL(q(\bold{W})\|p(\bold{W})) = \mathbb{E}_{q(\bold{W},\bold{z}_T)}[-KL(q(\bold{W}|\bold{z}_{T_f}) \| p(\bold{W})) + \log r(\bold{z}_{T_f} | \bold{W}) - \log q(\bold{z}_{T_f})]
%\end{equation}
%
%
\begin{align}\label{eq:lowerbound_loss}
    \mathcal{L}_{MNF.} &= -KL(q(\bold{W})\|p(\bold{W})) \nonumber \\
    &= \mathbb{E}_{q(\bold{W},\bold{z}_T)}[-KL(q(\bold{W}|\bold{z}_{T_f}) \| p(\bold{W})) + \log r(\bold{z}_{T_f} | \bold{W}) - \log q(\bold{z}_{T_f})]
\end{align}

The SELU activation functions, as presented by Klambauer et al. \cite{klambauer2017self}, are employed for their inherent self-normalizing properties, which ensure non-vanishing gradients. Their self-normalization nature could provide cases in which batch normalization is not needed, although this is data-dependent. We utilize SELU along with batch normalization to improve network convergence \cite{klambauer2017self}. We also note that \cite{ARRATIA2022166164} utilizes these activation functions.

\paragraph{Training}
Training and inference are performed utilizing a Python 3.9.12 environment with Pytorch 1.12.1 and CUDA 11.3. The model is trained for a maximum of 100 epochs, utilizing a batch size of 1028, in which training is stopped early if validation loss has plateaued. The model is trained using the \textit{Adam} optimizer with an initial learning rate of $5e^{-4}$, and is subject to a stepped learning rate function in which we decay $(\gamma)$ by an order of magnitude every 50 epochs (step size). It was found that decreasing the learning rate in such a fashion allows the network to converge to a more stable lower value. During training, it is important to correctly weight the KL loss contribution in such a way that it does not dominate, yet allows the convergence to informative posteriors. This too holds true in the relation between the physics informed and regression losses. Optimal values of $\alpha = 1.0$ and $\beta = 0.01$ were found through simple grid search optimization schemes.
The total dataset size is $\sim 12$ million events, split into a standard $70\%,15\%,15\%$ training, validation and testing split. This provides $\sim 2$ million events for testing purposes. Data injected to the network is scaled on the interval (-1,1), and targets are also scaled into the same interval. Table \ref{tab:training} provides the specs for training. 
\paragraph{Inference}
At inference we sample each individual event $10k$ times, taking the mean value as our final prediction. The epistemic uncertainty component is given by the standard deviation in these predictions. Each individual inference will also provide the corresponding aleatoric uncertainty, in which we again take the mean as our final aleatoric component. We perform this in batches of size 100, which corresponds to an overall batch of 1 million. This results in an event-level inference time of $20ms$. Note that this pipeline could be further vectorized to further decrease wall time.
Table \ref{tab:inference} provide the specs for inference.

\begin{table}[!]
\setlength{\tabcolsep}{50pt}
\centering
\begin{tabular}{ c  c } 
\hline
\textbf{Training Parameter} & \textbf{value} \\
\hline
Max Epochs & 100 \\
Batch Size & 1024 \\
Decay Steps & 50 \\
Decay Factor ($\gamma$) & 0.1 \\
Physics Loss Scale ($\alpha$) & 1.0 \\
KL Scale ($\beta$) & 0.01 \\
Training GPU Memory & $\sim 1$GB \\
Network memory on local storage & $\sim 7$MB \\
Trainable parameters & 611,247\\
Wall Time & $\sim$ 1 Day \\
\hline
\end{tabular}
 \caption{Training Specs of the ELUQuant architecture: training was performed with an Intel i7-12700k 12 Core CPU, Nvidia RTX 3090 24GB GPU and 64GB memory.}\label{tab:training}
\end{table}

\begin{table}[!]
\setlength{\tabcolsep}{50pt}
\centering
\begin{tabular}{ c  c } 
\hline
\textbf{Inference Parameter} & \textbf{value} \\
\hline
Number of Samples (N) & 10k \\
Batch Size & 100 \\
Inference GPU Memory & $\sim 24$GB \\
Inference Time per Event & $\sim 20ms$ \\
\hline
\end{tabular}
 \caption{Inference Specs of the ELUQuant architecture. Same computing resources of Table \ref{tab:training}.}\label{tab:inference}
\end{table}
% inference was performed with an Intel i7-12700k 12 Core CPU, Nvidia RTX 3090 24GB GPU and 64GB memory

\paragraph{Limitations} 

Identifying a suitable network structure for a Bayesian network is generally not straightforward given the model is attempting to optimize a distribution of weights at each layer. %It is therefore important that the complexity of the network approaches a minimum to allow efficient convergence. In our studies, 
The problem has been alleviated by utilizing the network's deterministic counterpart (a DNN) to identify a minimal complexity model that provides acceptable performance. We then directly utilized the structure from this network in ELUQ.
Another aspect to consider, depending on the data,  during the initial stages of training is that 
%
%that the associated aleatoric uncertainty can be small in relation to the observable.
%During the initial training stages of the model 
%
 it is possible a poorly calibrated model produces poor regression targets with small aleatoric components, resulting in unstable fluctuations. In light of this, we bounded the minimum and maximum variances to improve training stability. The numerical values of the bounds were set such that they do not influence the learned aleatoric component, \textit{i.e.}, the network does not default to the minimum or maximum allowed value.

\section{Analysis and results}\label{sec:Results}

Our strategy began with training a streamlined DNN that, despite its reduced complexity compared to \cite{ARRATIA2022166164} (150k parameters compared to 1.2M), achieved similar performance. While ELUQ and the DNN share similar architectural layer sizes, ELUQ stands out by offering enhanced UQ not possible with a basic DNN. 
We utilized ELUQ to predict the DIS kinematic variables and their associated aleatoric and epistemic uncertainties. 
In what follows, we make comparisons with other traditional reconstruction methods, namely EL, JB, and DA, introduced in Sec. \ref{sec:DIS_UQ}. We will also incorporate DNN into the comparative visualizations. The section will be split into two parts. Sec.\ref{sec:gen_performance} will discuss the general performance of the architecture in relation to other methods, similar to what is done in \cite{ARRATIA2022166164}, and Sec.\ref{sec:unc_analysis} will provide detailed studies on the uncertainties produced by our model, and how their utilization can result in increased performance. 

\subsection{Regression Performance}\label{sec:gen_performance}

\begin{figure}[!]
    \centering
    \includegraphics[width=0.325\textwidth,trim={3.5cm 0 4.5cm 0},clip]{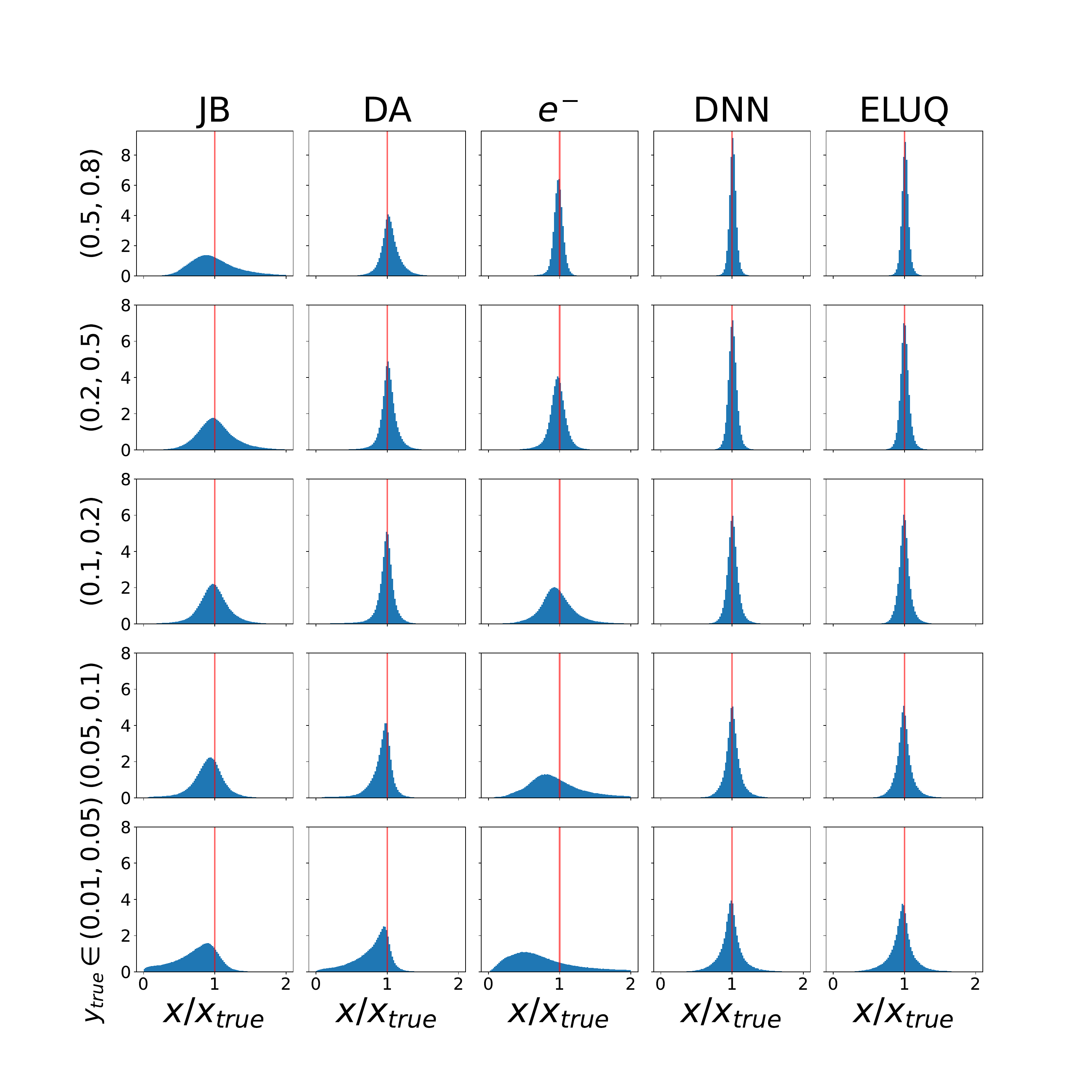} %
    \includegraphics[width=0.325\textwidth,trim={3.5cm 0 4.5cm 0},clip]{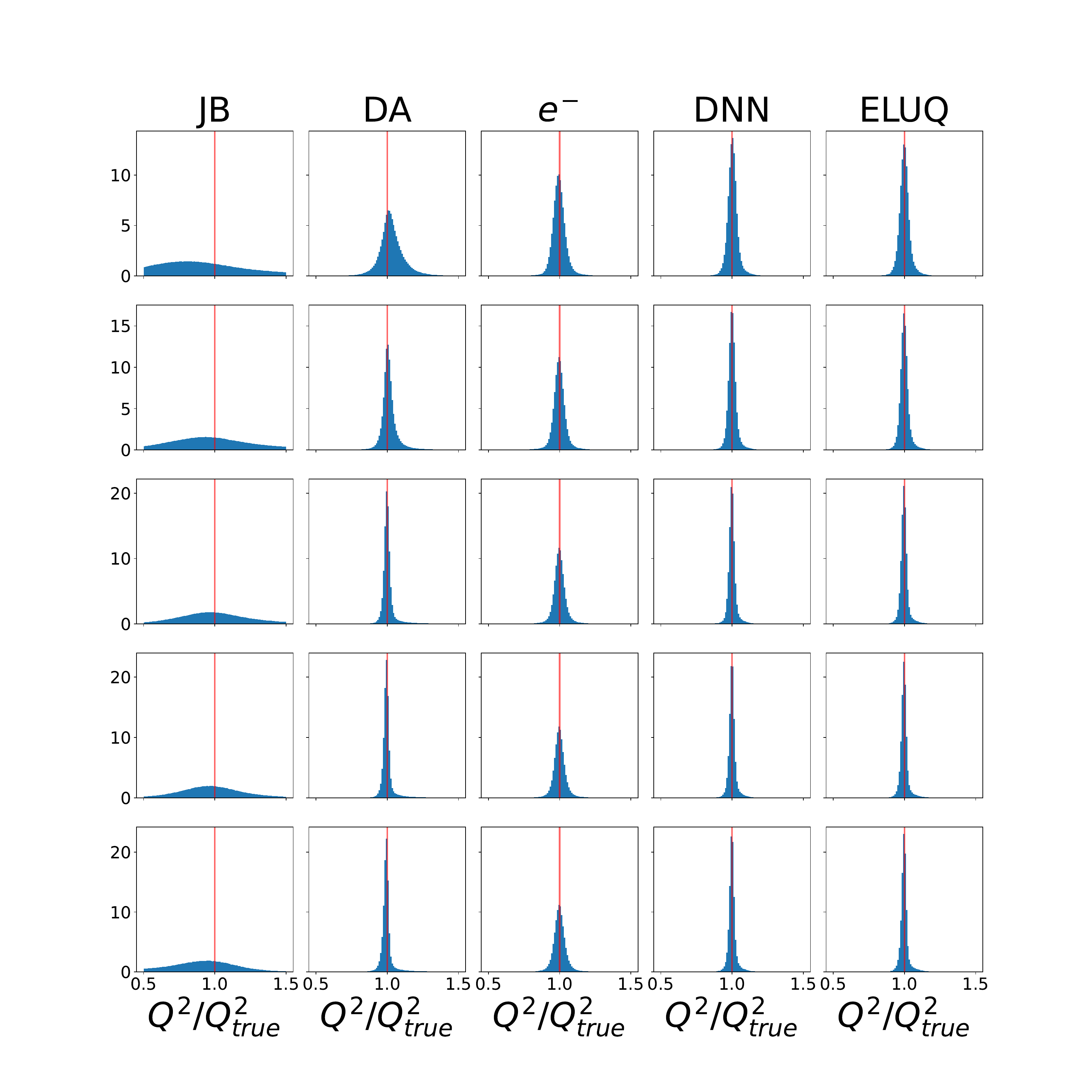} %
    \includegraphics[width=0.325\textwidth,trim={3.5cm 0 4.5cm 0},clip]{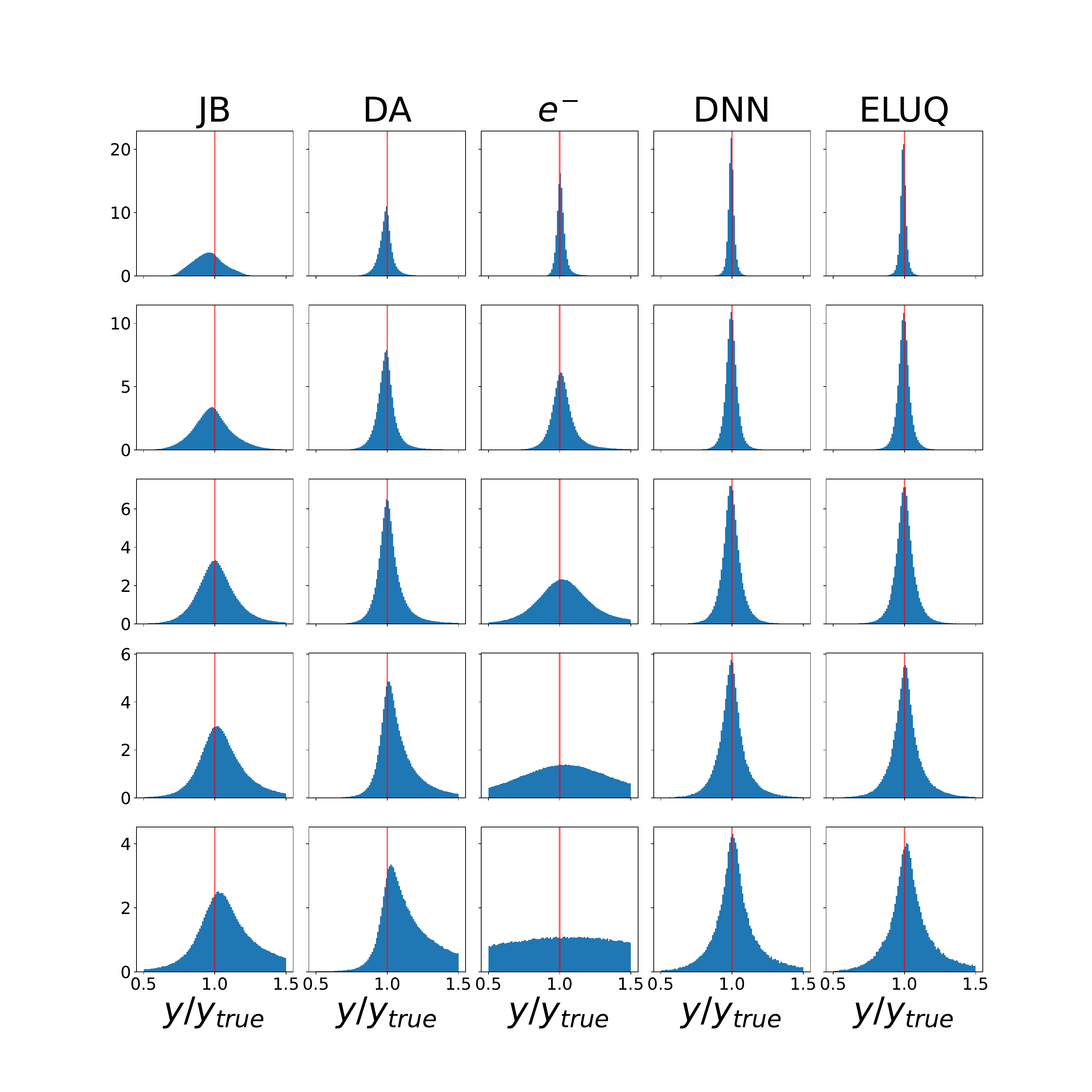}    
    \caption{Resolutions for $x$, $Q^{2}$, $y$ vs $y$ for different reconstruction methods.}
    \label{fig:resolutions_matrix}
\end{figure}

Fig. \ref{fig:resolutions_matrix} shows resolutions for $x$, $Q^{2}$, $y$ in bins of $y$ and comparison among the various reconstruction methods. 

We can immediately notice that the distributions of DNN and ELUQ look alike over the whole range in $y$ and for all the DIS kinematic variables. The choice of the binning in $y$ is to reproduce and compare with the results in \cite{ARRATIA2022166164}.
We also notice that DNN and ELUQ outperform traditional methods. As expected, the traditional methods do perform differently as a function of $y$: for example, the methods that mostly rely on the scattered electron yield the best resolution in events with large $y$, but their resolution on $x$ quickly diverges at low $y$. 
As already discussed, with ELUQ we can calculate uncertainties at the event-level. Given an observable \(\hat{O}_k\) for the \(k^{\text{th}}\) event and its associated uncertainty \(\sigma_k\), the weighted average of the observable over the entire dataset using uncertainty level information, and its associated uncertainty, are given by:
\begin{align}\label{eq:weighted_average}
\langle\hat{O}\rangle_{w} = \frac{\sum_{k=1}^{N} \frac{\hat{O}_{k}}{\sigma^{2}_{k}}}{\sum_{k=1}^{N} \frac{1}{\sigma^{2}_{k}}}, \  \
 \sigma_{w}(\langle\hat{O}\rangle_{w}) = \frac{1}{\sqrt{\sum_{k=1}^{N} \frac{1}{\sigma^{2}_{k}}}}.
\end{align}
While other methods do not provide direct access to event-level uncertainty, comparisons between methods are still feasible. To facilitate this, the expected event-level uncertainty is approximated using the RMS as detailed in \cite{ARRATIA2022166164}. The RMS is then compared to the event-level equivalent derived from the weighted uncertainty, given by \(\approx \sigma_{w} \cdot \sqrt{N}\). Notice that due to the large statistics, the uncertainty on the averages will be exceedingly small and may be challenging to visually discern otherwise.

\begin{figure}[!]
    \centering
    \includegraphics[width=0.32\textwidth]{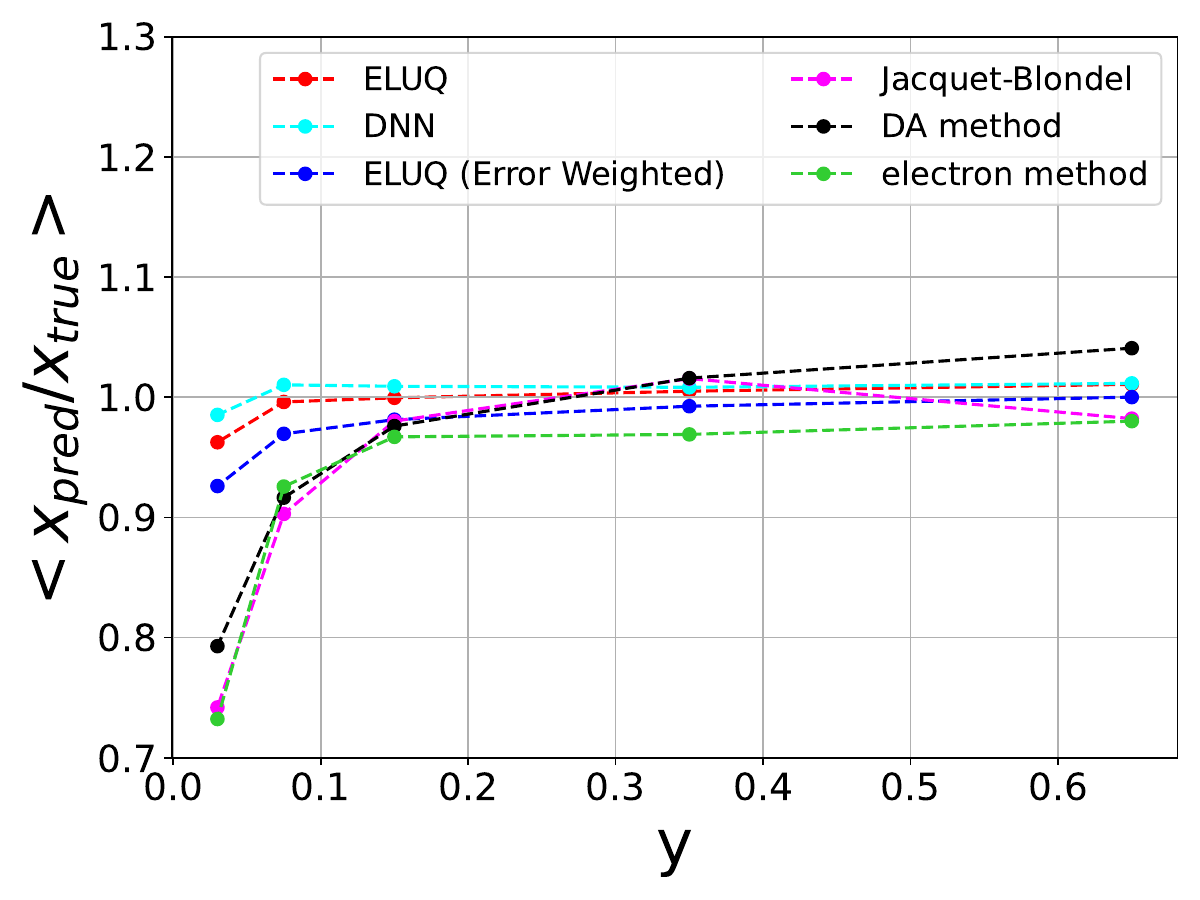} %
    \includegraphics[width=0.32\textwidth]{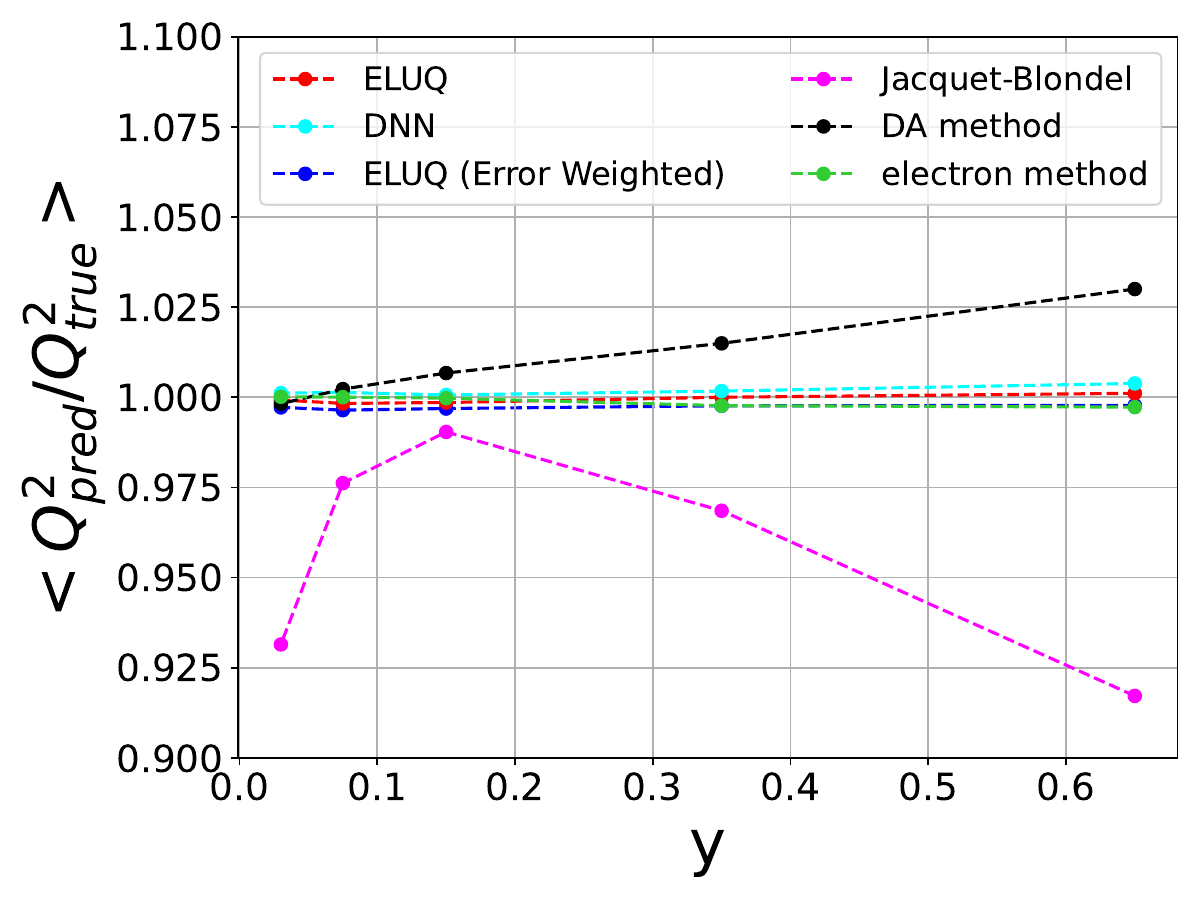} %
    \includegraphics[width=0.32\textwidth]{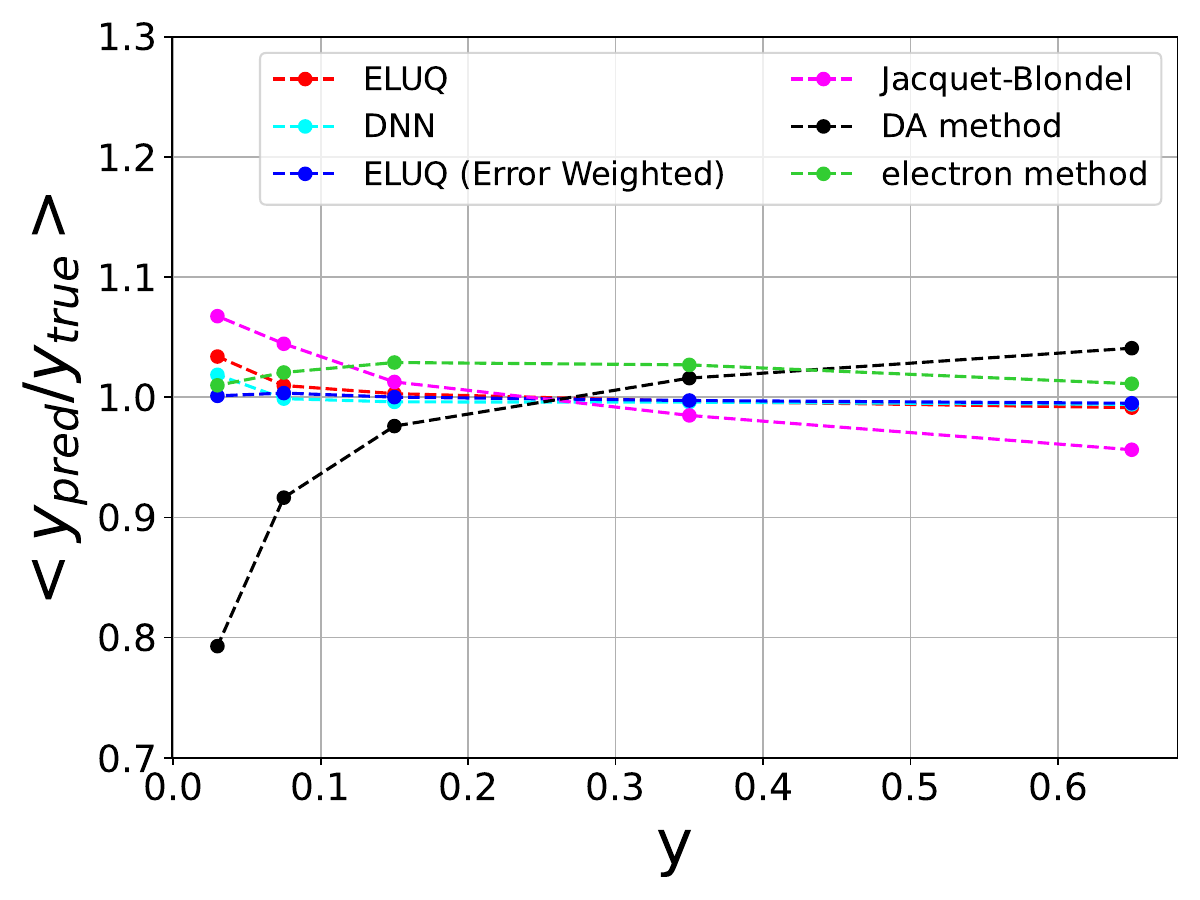} 
    \caption{The average predicted-to-truth ratio of the DIS kinematics for ELUQ and DNN compared to the classical methods, in bins of $y$.}
    \label{fig:ratio_vs_y_all}
\end{figure}

\begin{figure}[!]
    \centering
    \includegraphics[width=0.32\textwidth]{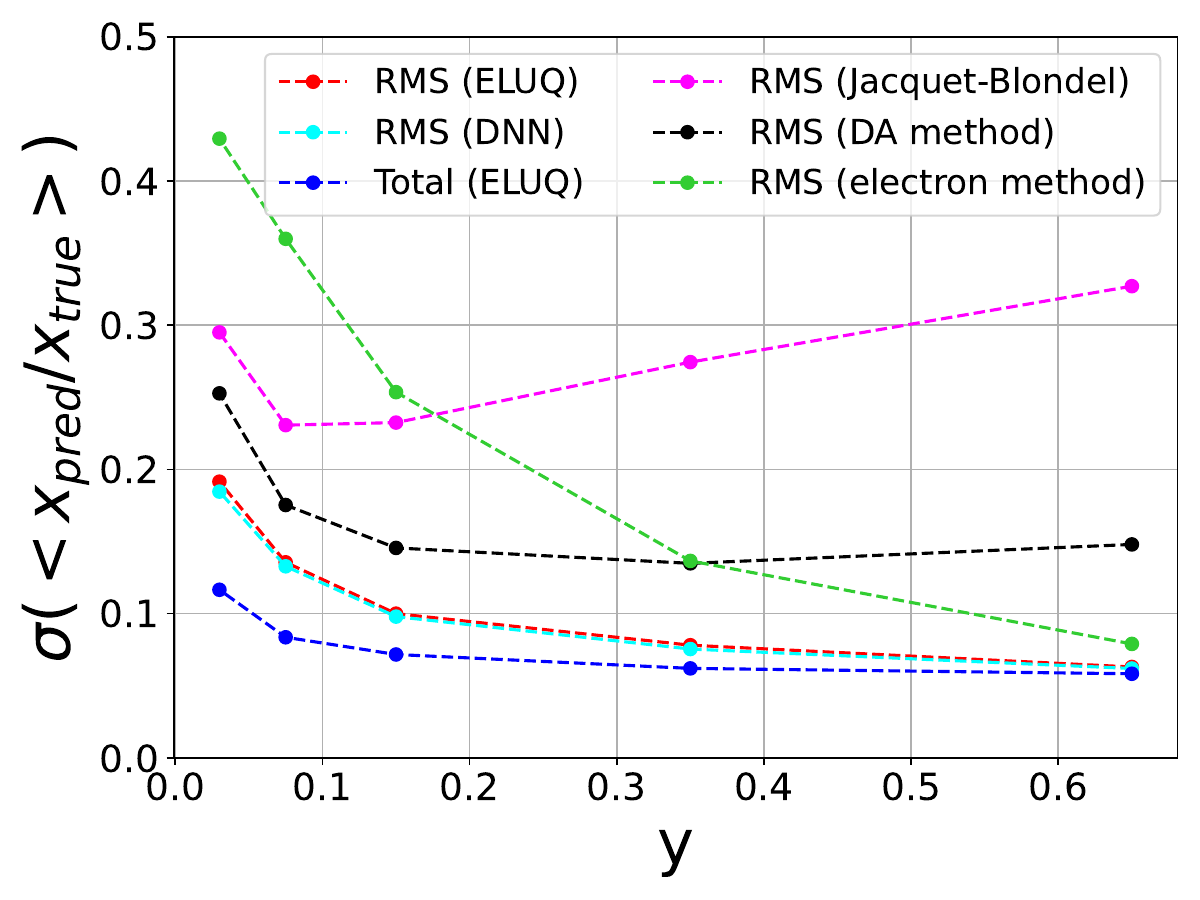} %
    \includegraphics[width=0.32\textwidth]{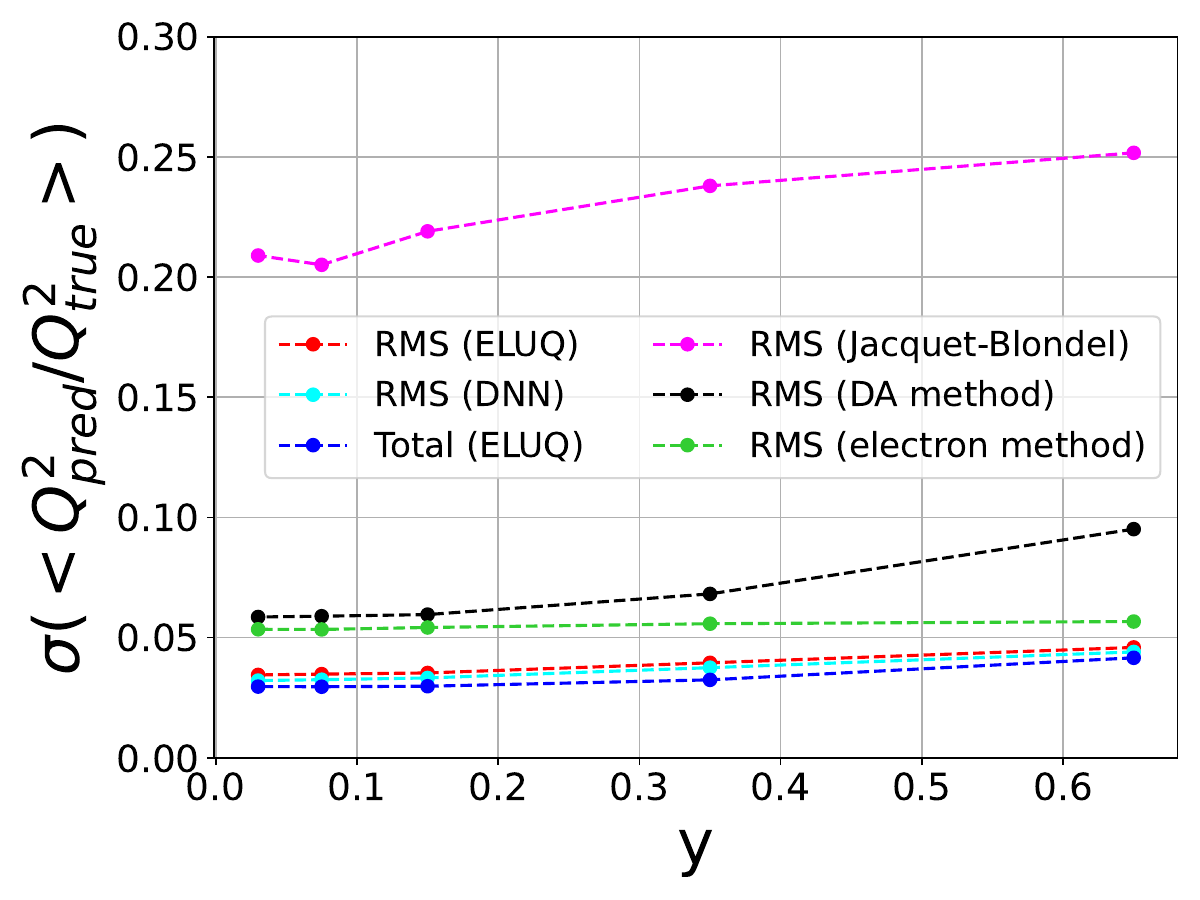} 
    \includegraphics[width=0.32\textwidth]{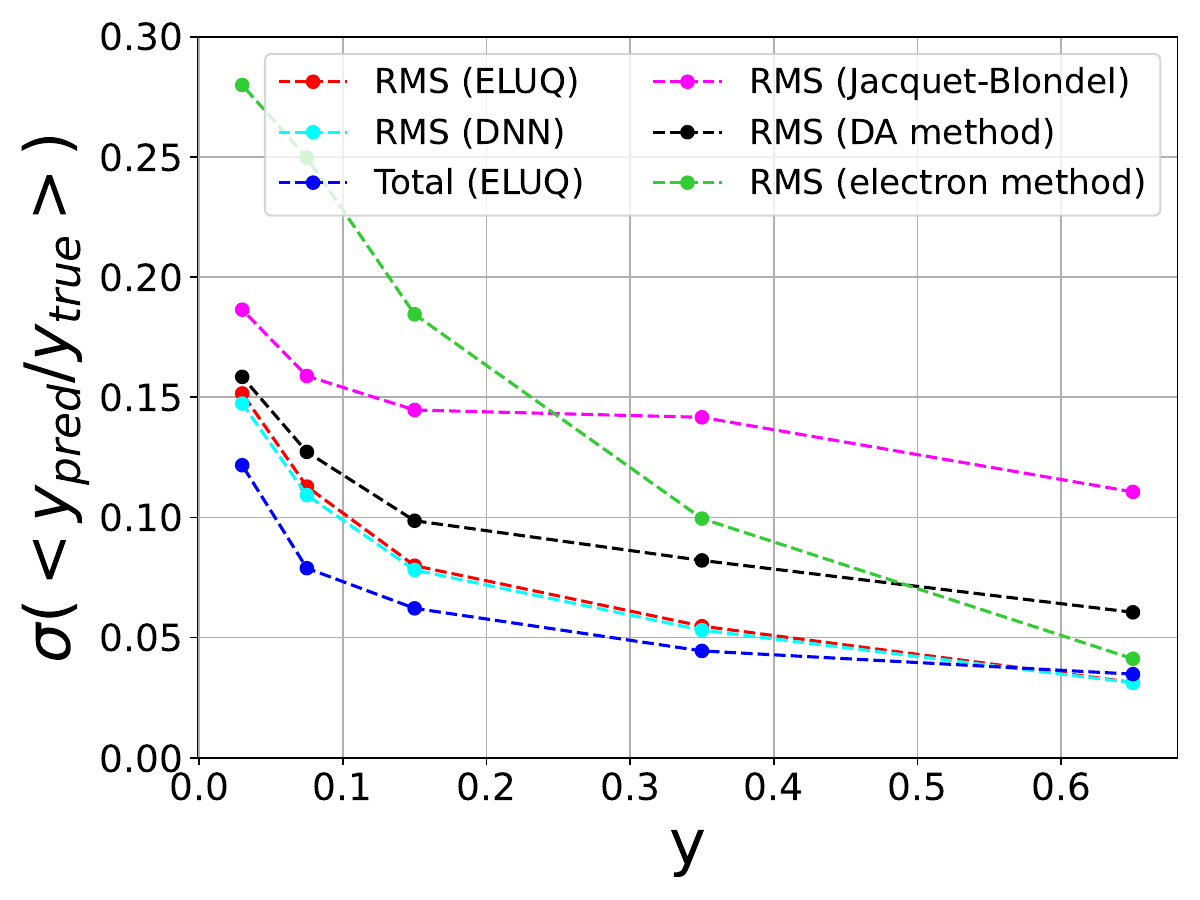} 
    \caption{Average event-level uncertainty on the predicted-to-true ratio in bins of $y$ for ELUQ, compared to the RMS from DNN and from classical methods.}
    \label{fig:unc_vs_y_all}
\end{figure} 

Fig. \ref{fig:ratio_vs_y_all} and Fig. \ref{fig:unc_vs_y_all} show the (weighted) average ratio of the predicted observables normalized to their ground truth, $<R_{O}> = <\hat{O}_{pred.}/\hat{O}_{true}>$, and the event-level uncertainties, respectively, in bins of the inelasticity $y$.

Fig. \ref{fig:ratio_vs_y_all} shows a drop in the $<R_{x}>$ at low $y$, where the RMS resolution for $y$ and $x$ increase, even for the DNN and ELUQ reconstruction, as shown in Fig. \ref{fig:unc_vs_y_all}. 
According to \cite{ARRATIA2022166164}, this results may be attributed to further acceptance, noise, or resolution effects that deteriorate the measurement of the HFS. Notice that the weighted average is slightly more affected by this flaw in reconstruction performance than the arithmetic average.

\begin{figure}[!]
    \centering
    \includegraphics[width=0.32\textwidth]{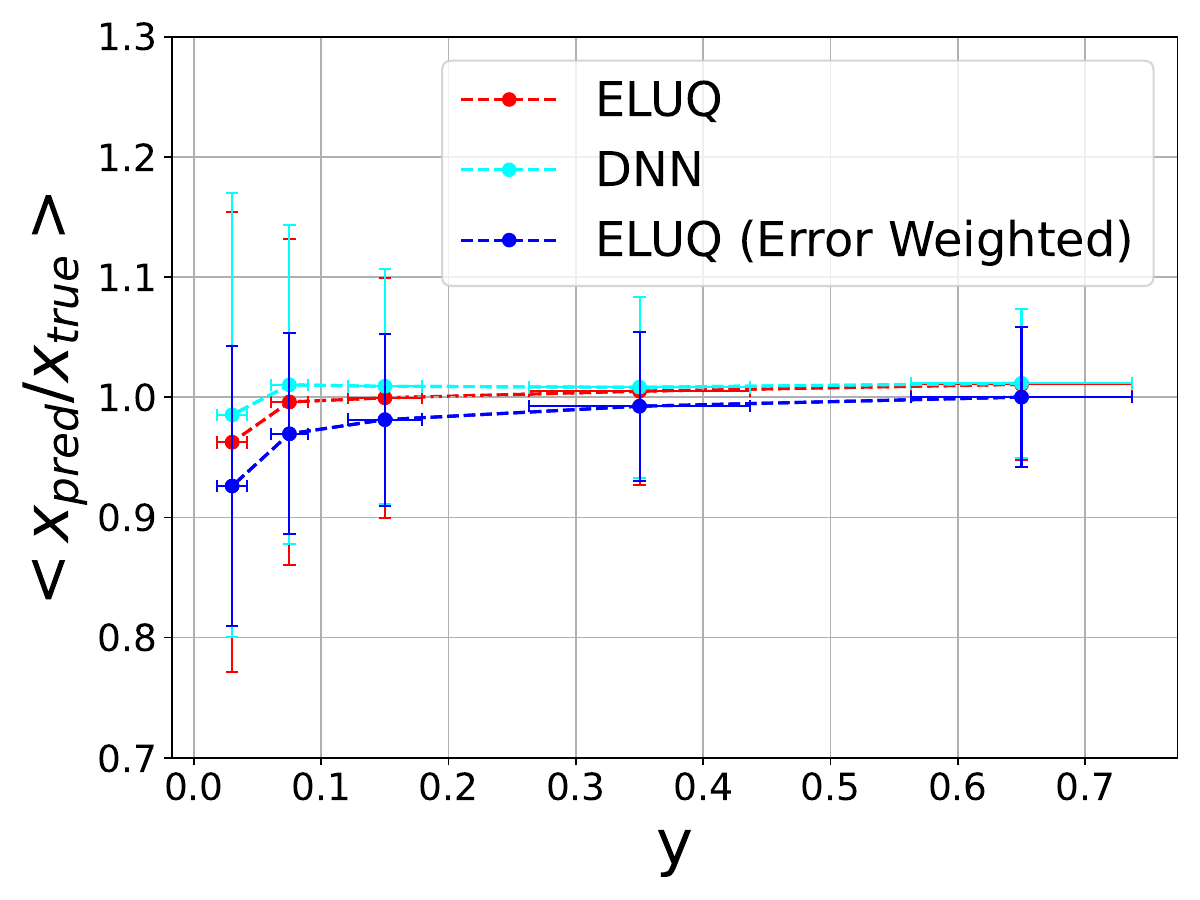} %
    \includegraphics[width=0.32\textwidth]{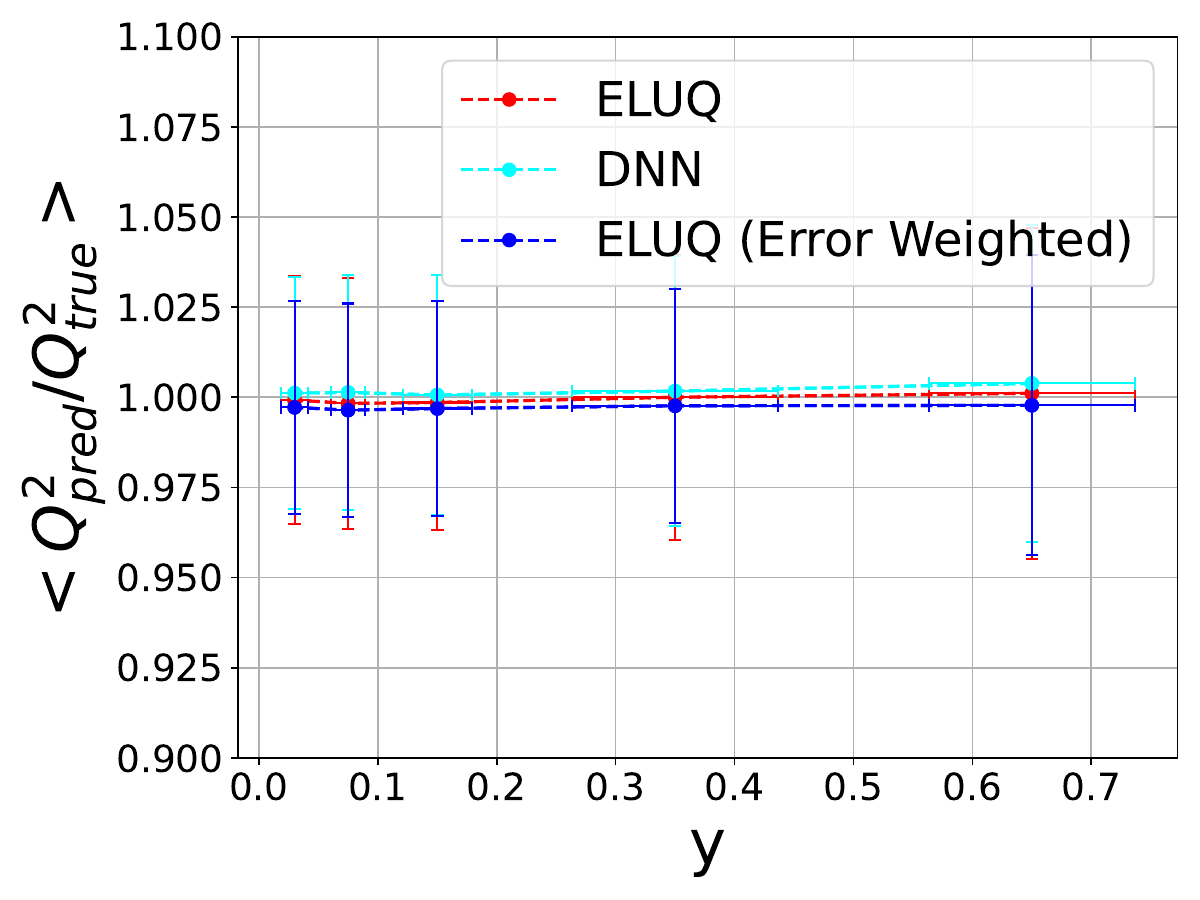} 
    \includegraphics[width=0.32\textwidth]{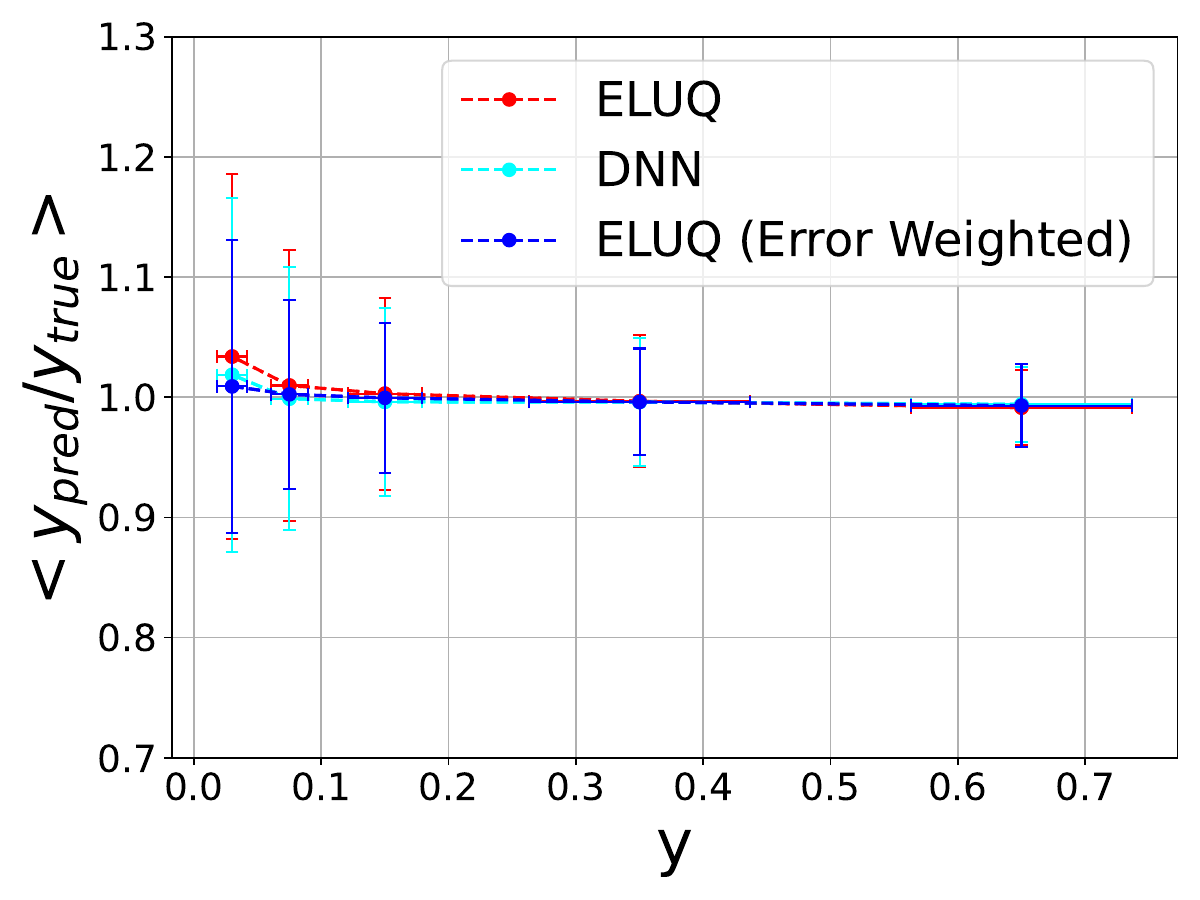}
    \caption{Average predicted-to-true ratio in bins of $y$ for ELUQ, DNN with event-level uncertainty. In blue we also show for ELUQ the smaller event-level uncertainty on the weighted average compared to the RMS on ELUQ and DNN.}
    \label{fig:ELUQ_vs_DNN}
\end{figure}

Fig. \ref{fig:ELUQ_vs_DNN} shows a comparison of the ratios between ELUQ and DNN; for ELUQ, we report both the RMS and the event-level equivalent weighted uncertainty. 
Notice that the total uncertainty at the event-level for ELUQ is given by the sum in quadrature of the aleatoric and epsitemic components, $i.e.$, $\sigma_{tot} = \sigma_{ale} \oplus \sigma_{epi}$. 

\subsection{Uncertainty Analysis}\label{sec:unc_analysis}
The validation criteria of our model are two-fold. In the previous section, we validated regression performance in comparison to the model's deterministic counterpart (DNN), both with and without the inclusion of information from uncertainties. Showing the benefits of access to event-level uncertainty in relation to performance.
In what follows we validate the event-level uncertainty components individually. We conduct a series of closure tests on the aleatoric component to show the event-level quantities propagate correctly at the histogram level. %in comparison to the RMS. 
We also conduct closure tests on the epistemic component in which we show the uncertainty generated by our model decreases as a function of model calibration.

\begin{figure}[!]
    \centering
    \includegraphics[width=0.32\textwidth]{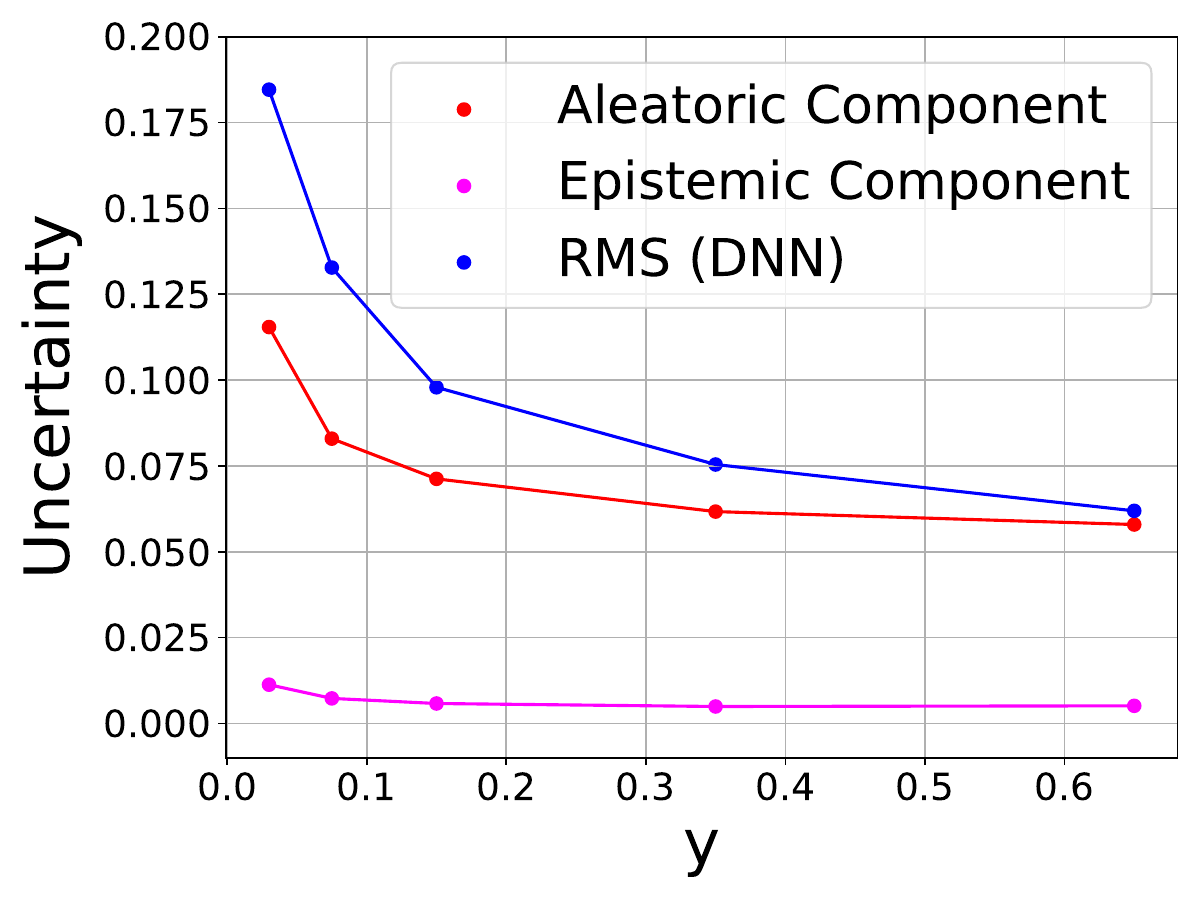} %
    \includegraphics[width=0.32\textwidth]{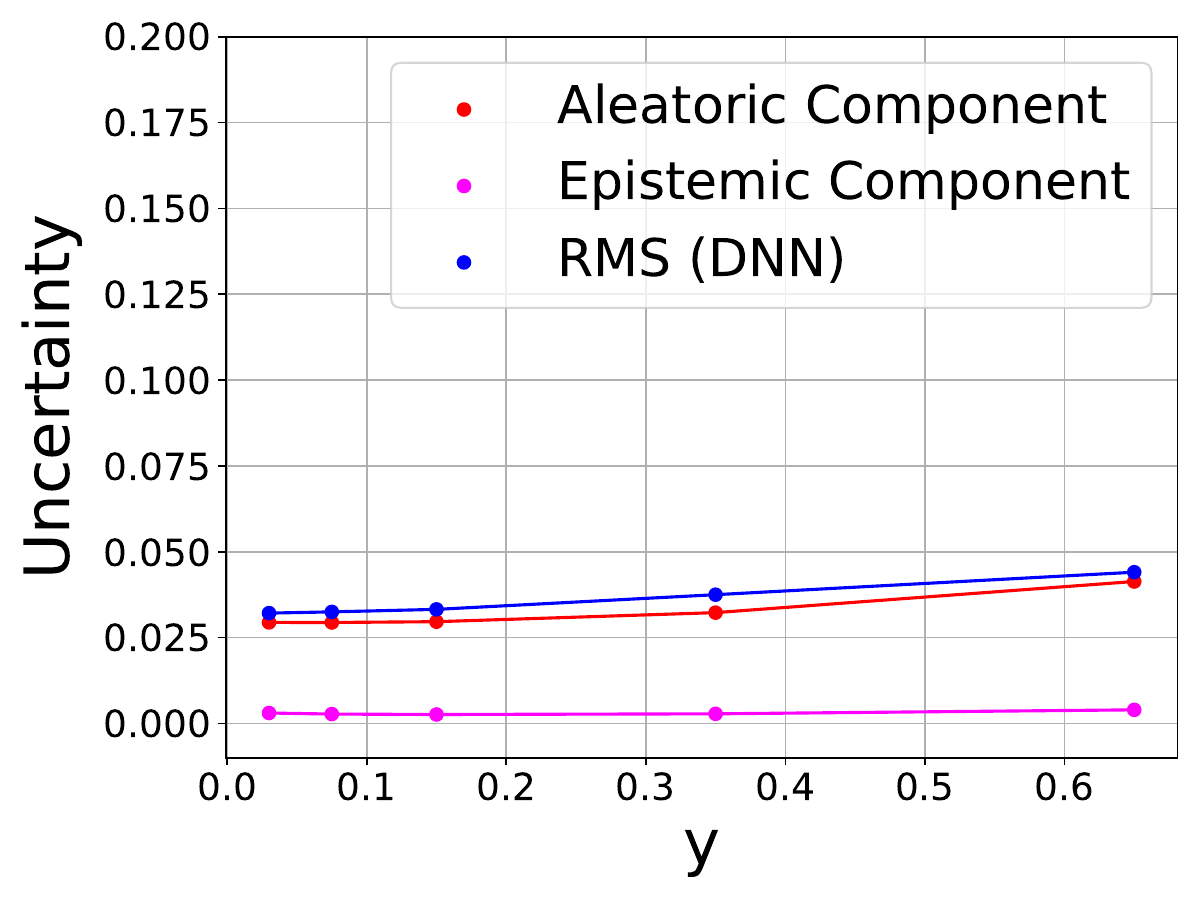} %
    \includegraphics[width=0.32\textwidth]{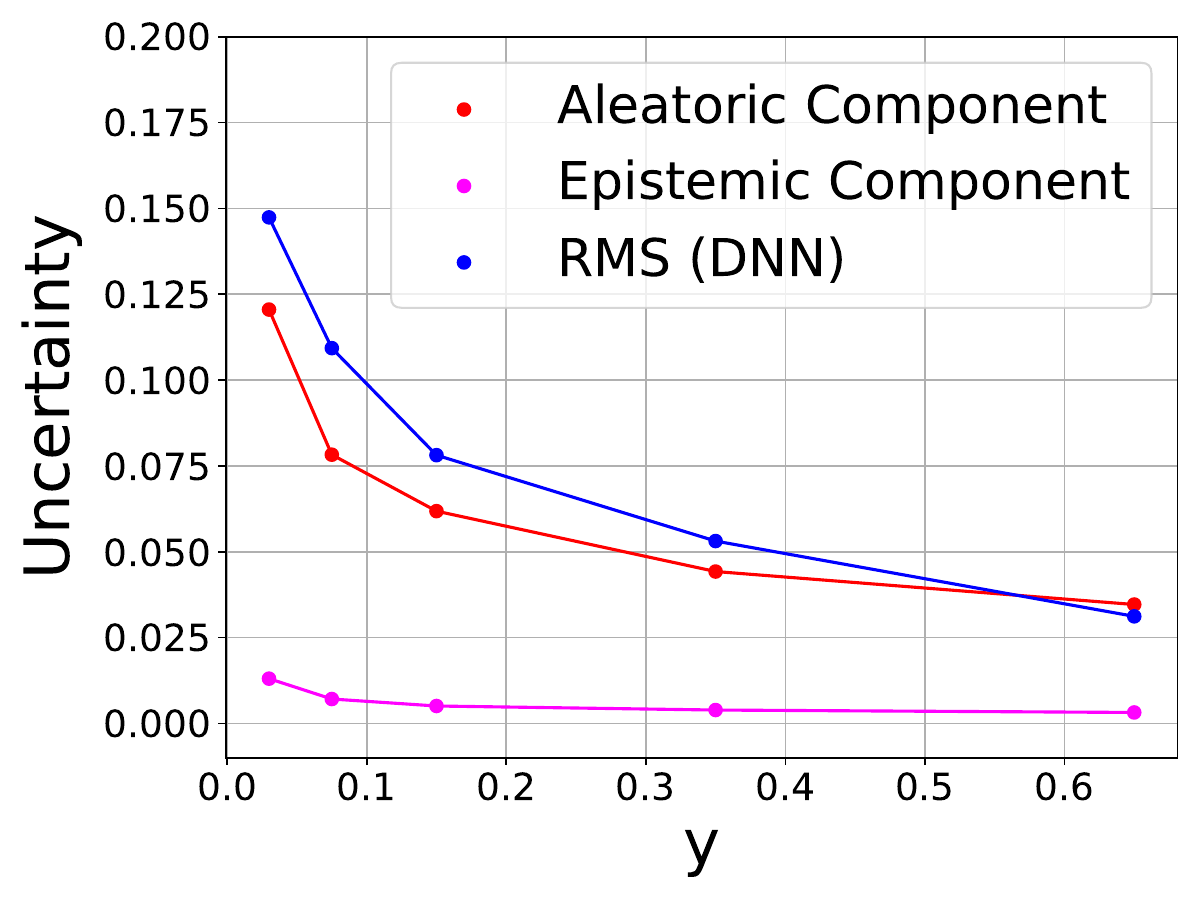} \\
    \caption{Average event-level aleatoric and epistemic uncertainty on the reconstruction-to-true ratio for $x$ (left), $Q^2$ (middle) and $y$ (right). ELUQ, compared to the RMS obtained from the DNN model.}
    \label{fig:uncertainty_rms_comparison}
\end{figure}

\begin{figure}[!]
    \centering
    \includegraphics[width=\textwidth]{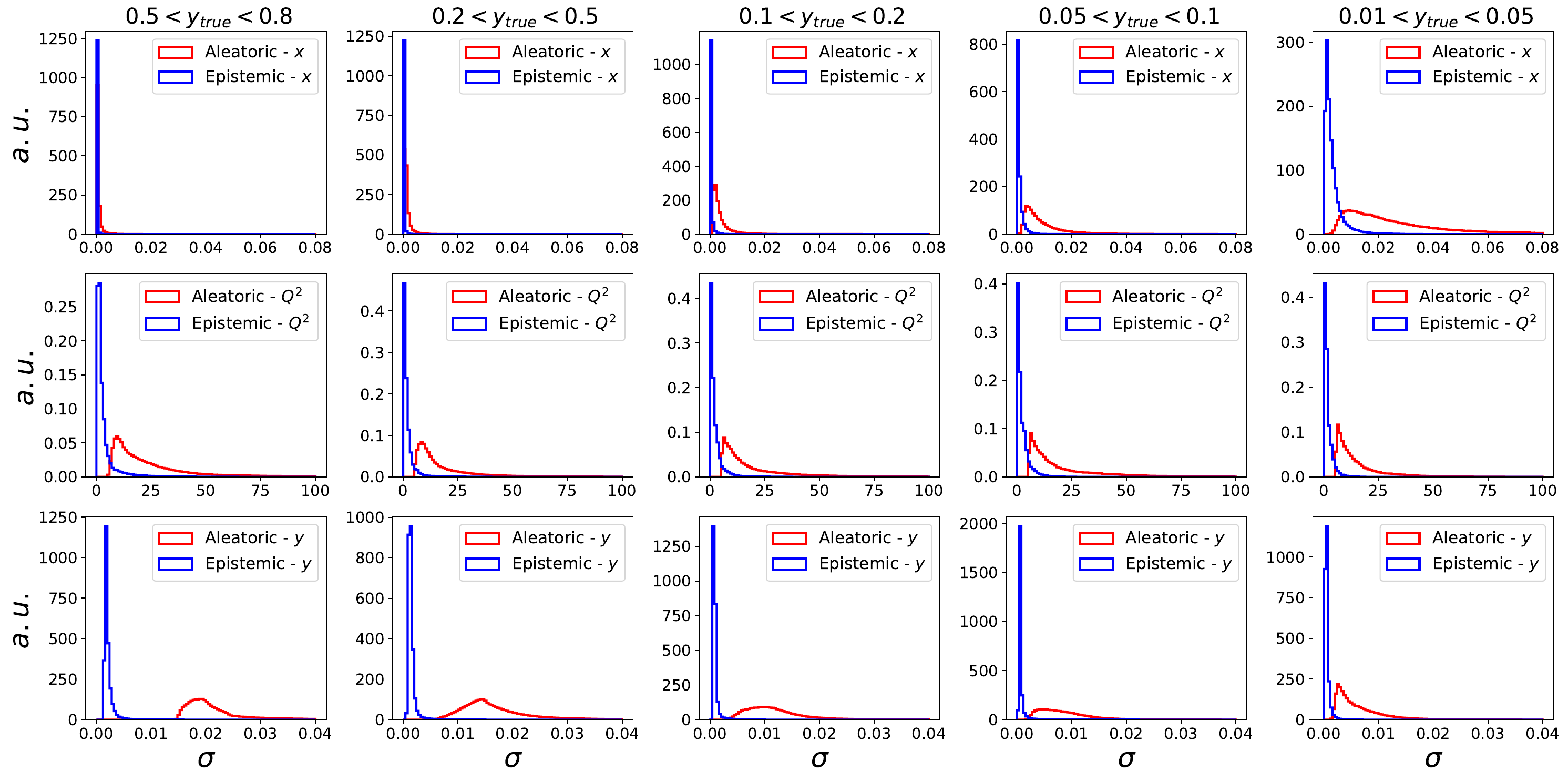} %
    \caption{Histograms of the aleatoric and epistemic uncertainties in bins of $y$ for $x$, $Q^{2}$, $y$.}
    \label{fig:ale_epi_hists}
\end{figure}

Fig. \ref{fig:uncertainty_rms_comparison} shows a comparison between $\sigma_{ale}$, $\sigma_{epi}$ and the RMS from DNN, for the three regressed variates $x$, (left), $Q^2$ (middle), $y$ (right).
Fig. \ref{fig:ale_epi_hists} presents a detailed analysis of the histograms representing event-level occurrences of \(\sigma_{ale}\) and \(\sigma_{epi}\) uncertainties on \(x\), \(Q^{2}\), and \(y\). These uncertainties are examined in bins of \(y\).

\begin{figure}[!]
    \centering
    \includegraphics[width=0.32\textwidth]{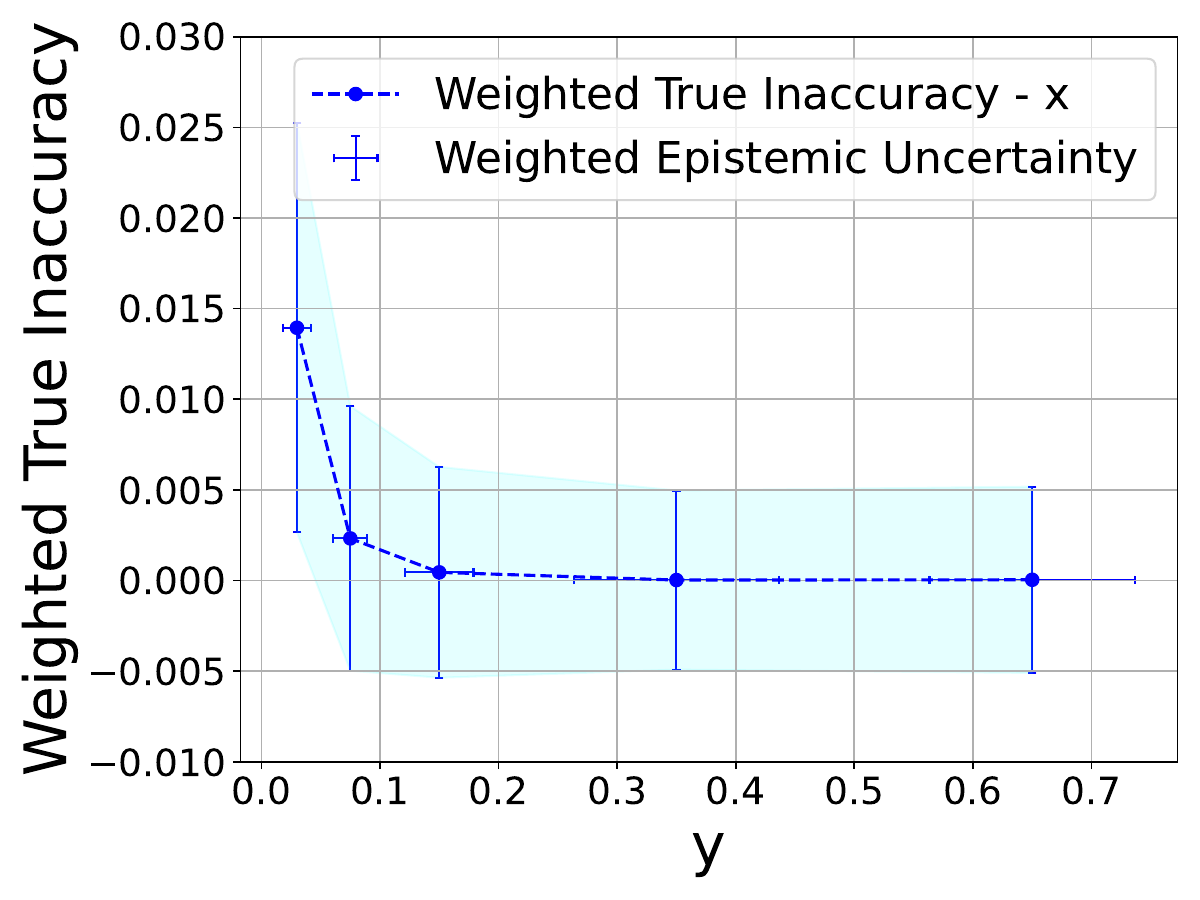} %
    \includegraphics[width=0.32\textwidth]{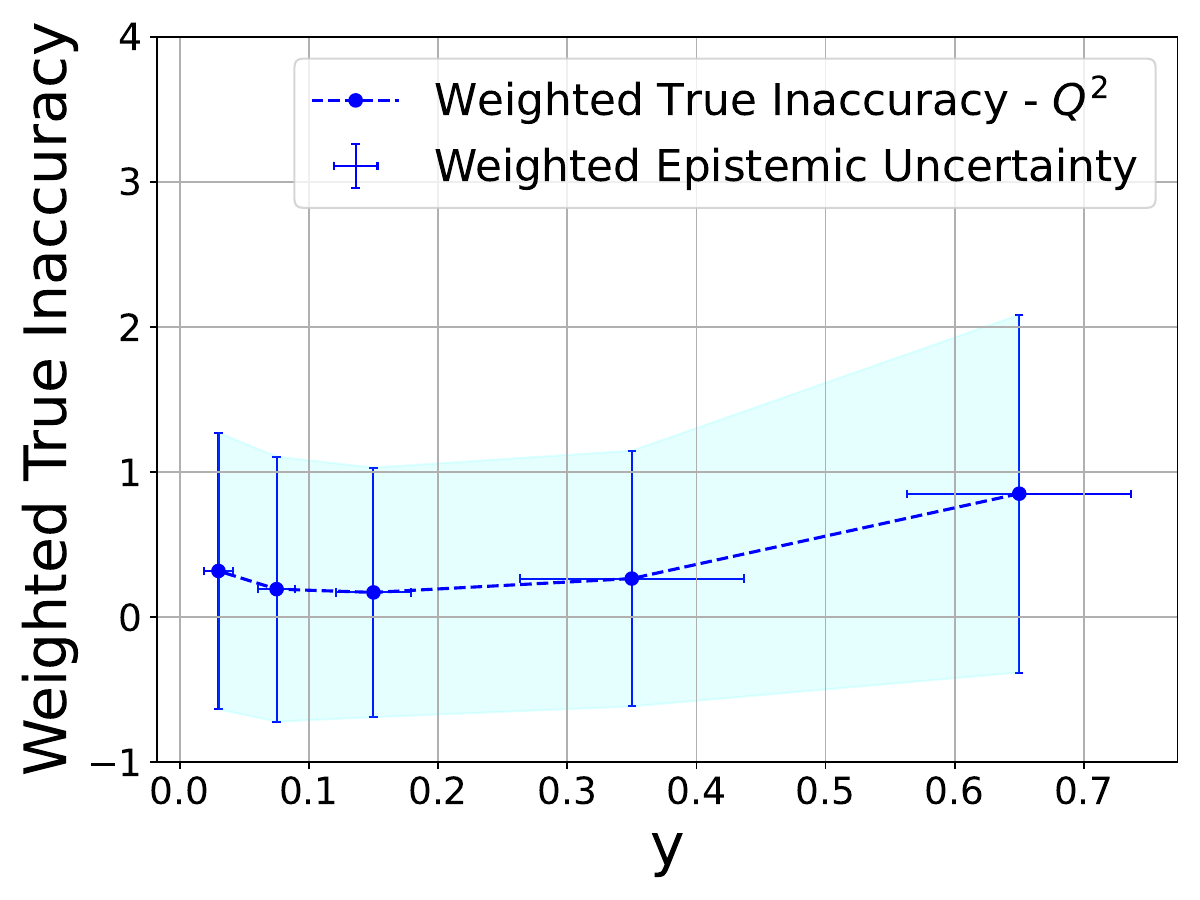} %
    \includegraphics[width=0.32\textwidth]{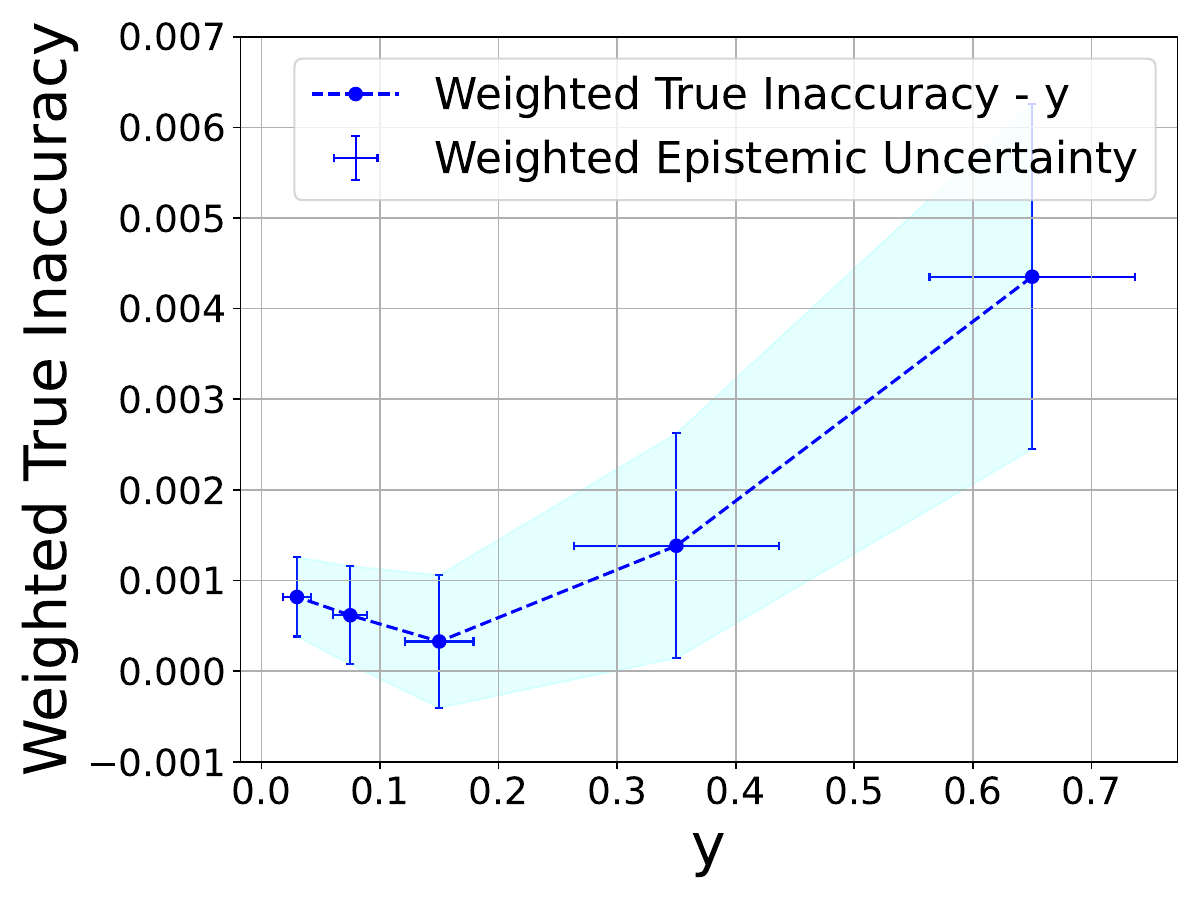} 
    \caption{Average event-level epistemic vs weighted true inaccuracy for the DIS kinematics in bins of $y$. The epistemic increases with the inaccuracy.}
    \label{fig:epistemic_vs_inaccuracy}
\end{figure}

\begin{figure}[!]
    \centering
    \includegraphics[width=0.32\textwidth]{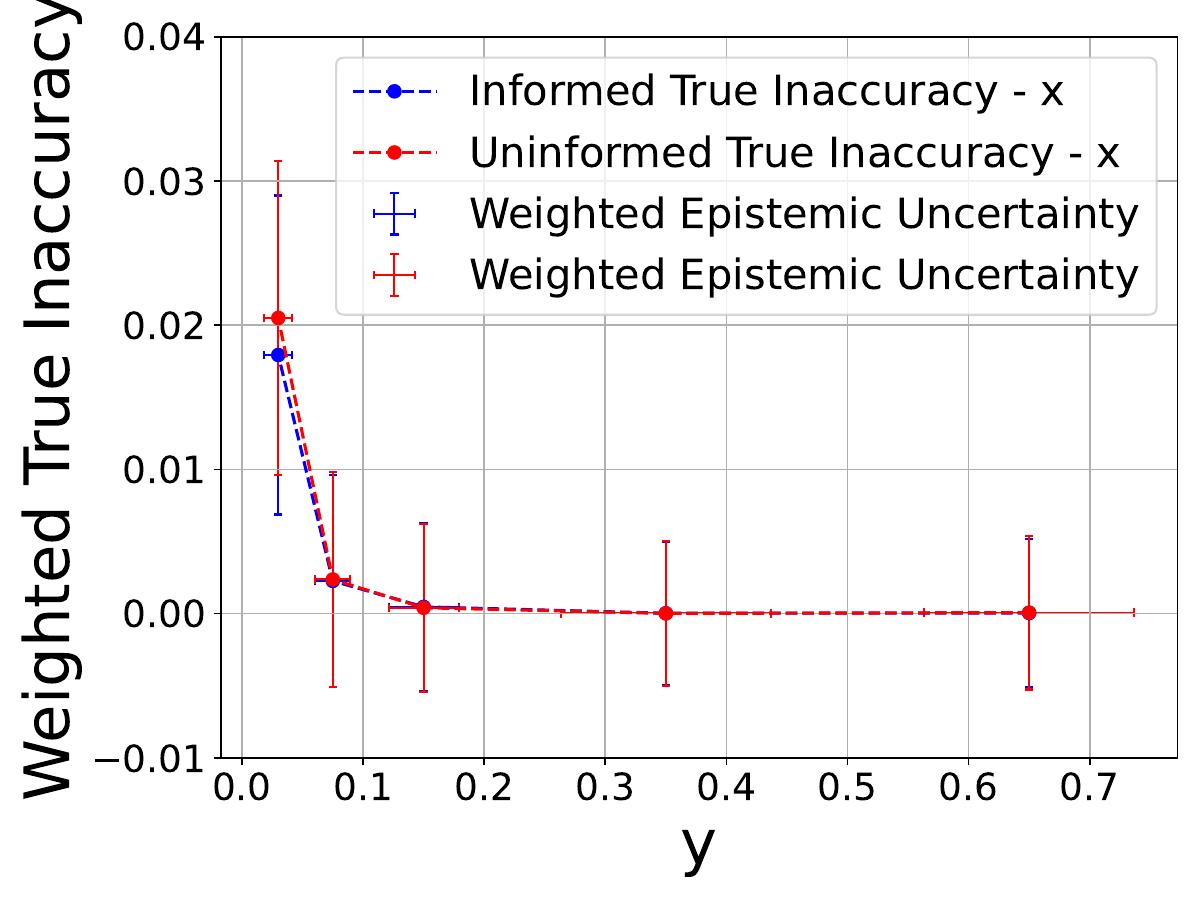} %
    \includegraphics[width=0.32\textwidth]{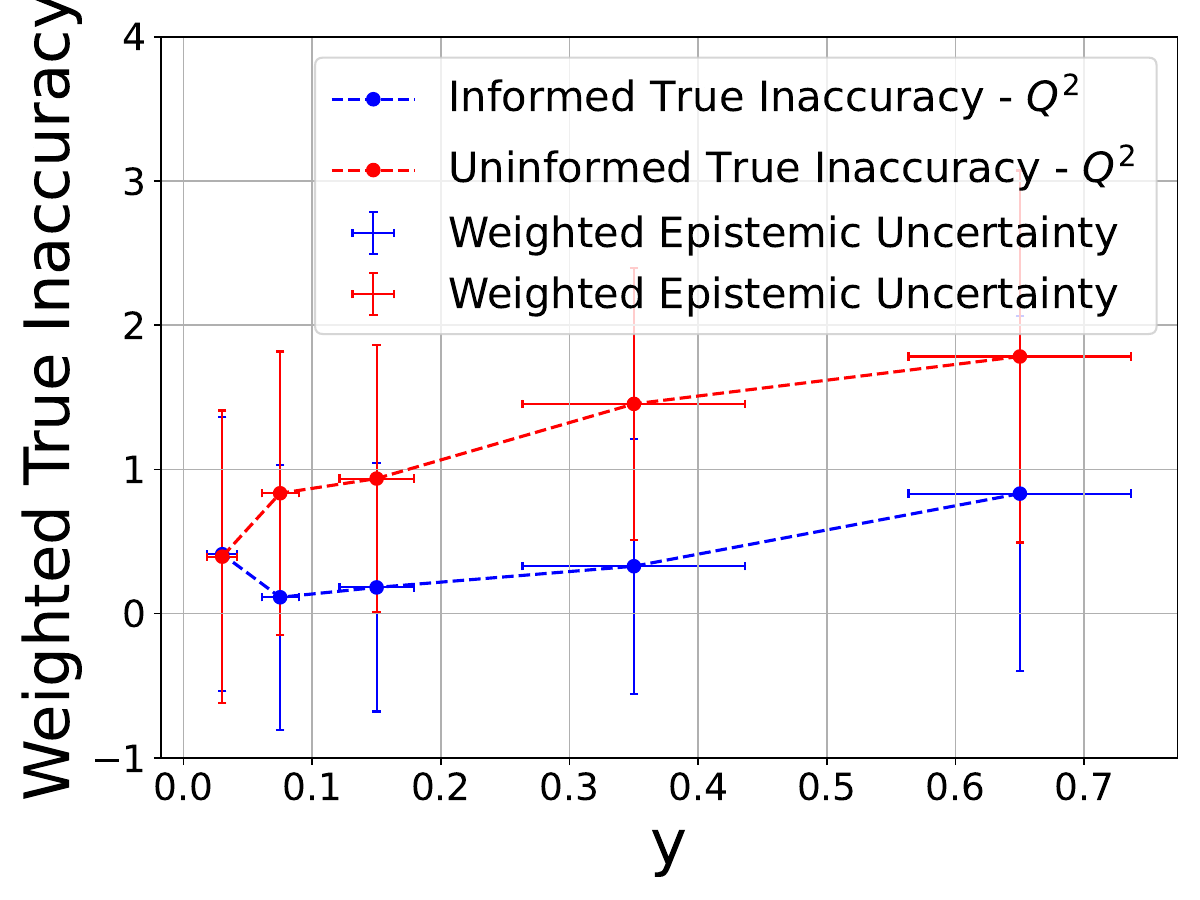} 
    \includegraphics[width=0.32\textwidth]{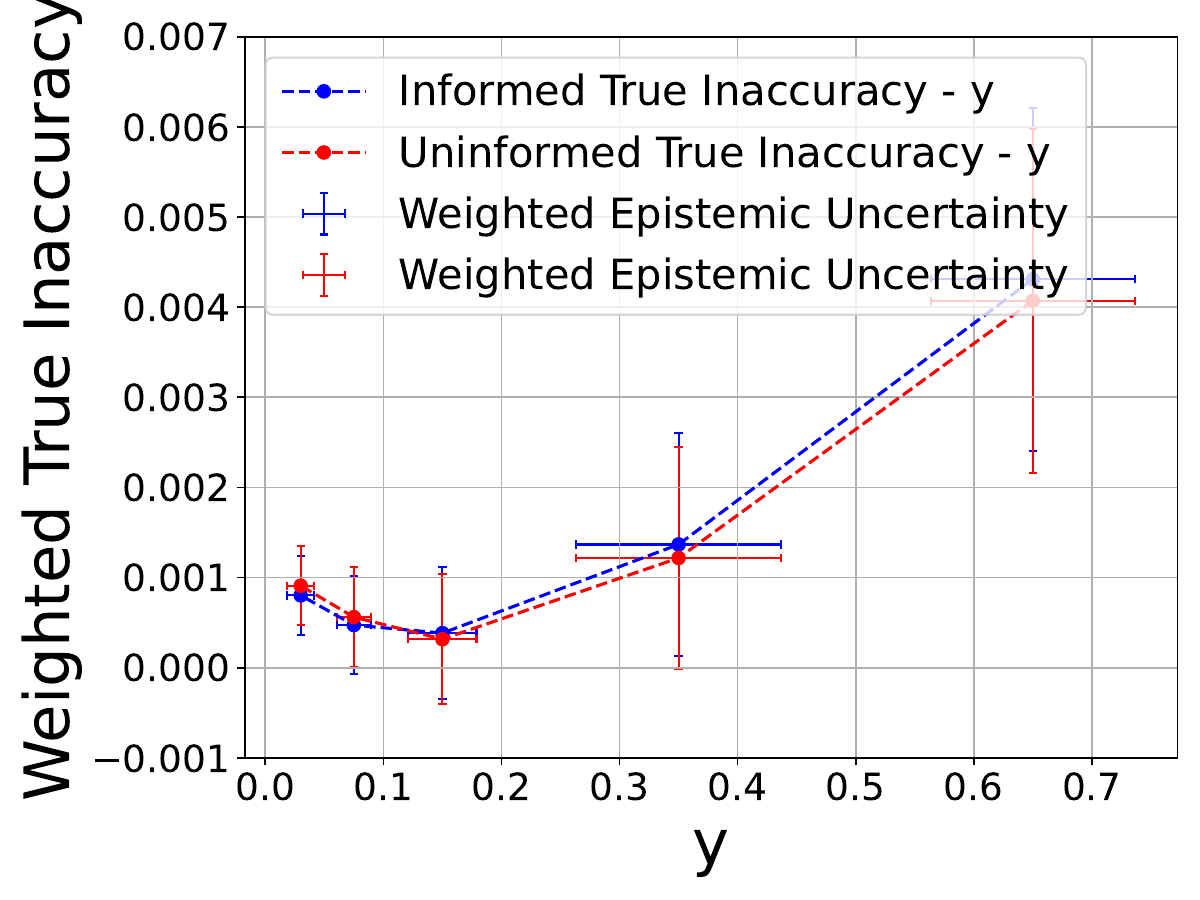} 
    \caption{Epistemic uncertainty vs true inaccuracy in bins of $y$, for $x$, $Q^{2}$, $y$. The plots show ELUQ trained with and w/o the physics-informed loss, its inclusion providing a lower inaccuracy for $Q^{2}$.}
    \label{fig:inaccuracy_vs_physics}
\end{figure}

Closure tests support the reliability of the aleatoric and epistemic uncertainties extracted. For instance, as shown in Table \ref{tab:aleatoric}, aleatoric uncertainties are consistent with the RMS of a DNN in bins of $y$ (visually depicted in Fig. \ref{fig:uncertainty_rms_comparison}) where the regressed observables manifest as Gaussian distributions
not affected by inaccuracy, that is, centered at the expected mean from ground truth. Notably, epistemic
uncertainty—originating from the same multivariate normal distribution characterizing the aleatoric term in the loss function—amplifies in response to increased inaccuracy with respect to ground truth, see Fig. \ref{fig:epistemic_vs_inaccuracy}.
\begin{table}[!]
    \centering
    \resizebox{0.975\textwidth}{!}{%
    \begin{tabular}{c c c c c c c c c c c c c c}
    \toprule
    y bin  & RMS(x$_{DA}$) & RMS(x$_{ele}$) & RMS(x$_{DNN}$)  & $\sigma$(x) & RMS(Q$^{2}_{DA}$) & RMS(Q$^{2}_{ele}$) & RMS(Q$^{2}_{DNN}$)  & $\sigma$(Q$^{2}$) & RMS(y$_{DA}$) & RMS(y$_{ele}$) & RMS(y$_{DNN}$)  & $\sigma$(y)\\
    \midrule
    \textbf{(0.5, 0.8)} & 0.15 & 0.079 & 0.062 & \textbf{0.058} & 0.095 & 0.057 & 0.044 & \textbf{0.041} & 0.061 & 0.041 & 0.031 & \textbf{0.035} \\
    \textbf{(0.2, 0.5)} & 0.13 & 0.14 & 0.075 & \textbf{0.062} & 0.068 & 0.056 & 0.038 & \textbf{0.032} & 0.082 & 0.100 & 0.053 & \textbf{0.044} \\
    \textbf{(0.1, 0.2)} & 0.15 & 0.25 & 0.098 & \textbf{0.071} & 0.060 & 0.054 & 0.033 & \textbf{0.030} & 0.099 & 0.18 & 0.078 & \textbf{0.062} \\
    \textbf{(0.05, 0.1)} & 0.18 & 0.36 & 0.13 & \textbf{0.083} & 0.059 & 0.053 & 0.033 & \textbf{0.029}  & 0.13 & 0.25 & 0.11 & \textbf{0.078} \\
    \textbf{(0.01, 0.05)} & 0.25 & 0.43 & 0.18 & \textbf{0.12} & 0.059 & 0.053 & 0.032 & \textbf{0.029} & 0.16 & 0.28 & 0.15 & \textbf{0.12} \\
    \bottomrule
    \end{tabular}
    }
    \caption{Comparisons between the aleatoric uncertainty of ELUQ with the RMS of other methods, for the DIS kinematic variables $x$, $Q^{2}$, $y$.}
    \label{tab:aleatoric}
\end{table}
UQ studies have also been conducted to demonstrate the effect of the physics-informed term on the inaccuracy $|\bold{v}- \hat{\bold{v}}|$. Fig. \ref{fig:inaccuracy_vs_physics} shows that Eq. \eqref{eq:physics_loss} contributes to decrease the inaccuracy on $Q^{2}$; it also confirms that the epistemic increases if the inaccuracy gets larger.    

\begin{figure}[!]
    \centering
    \includegraphics[width=0.32\textwidth]{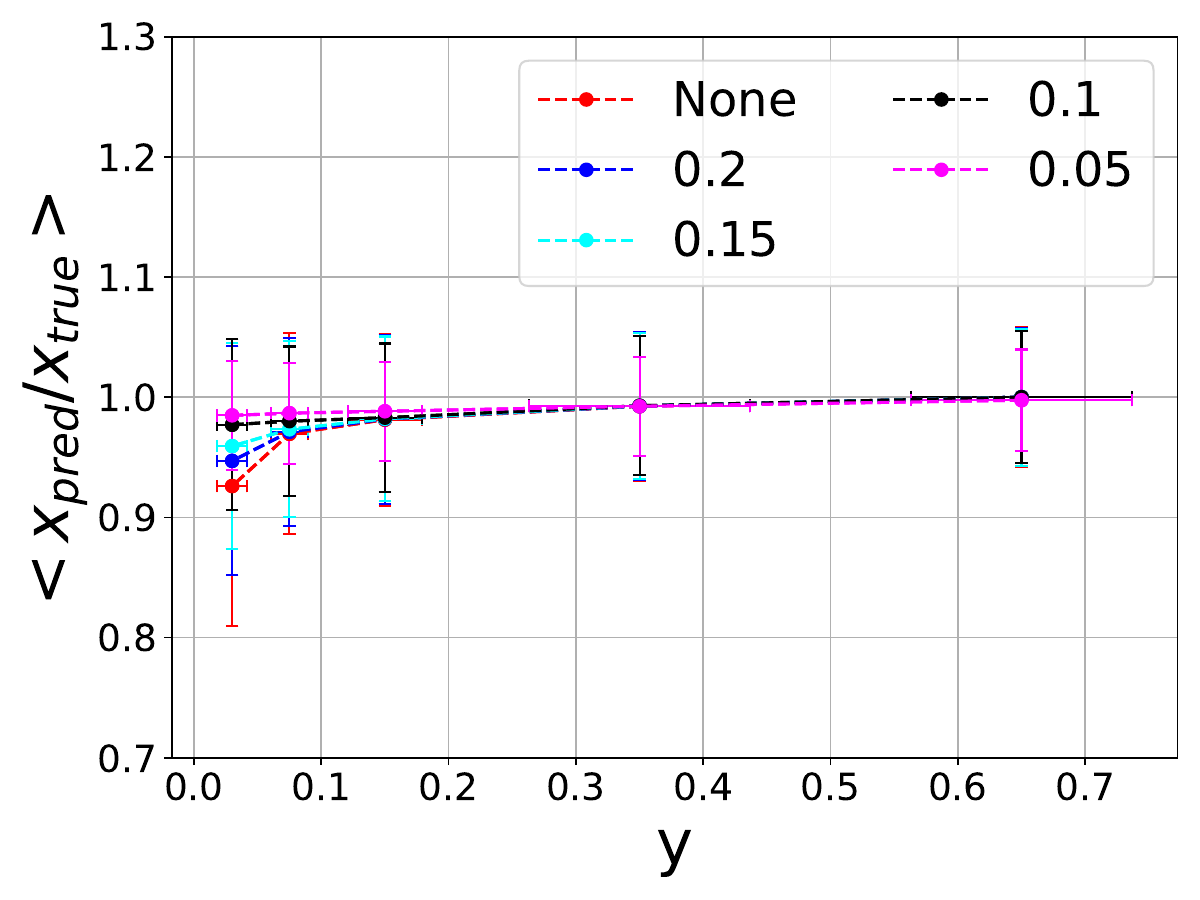} %
    \includegraphics[width=0.32\textwidth]{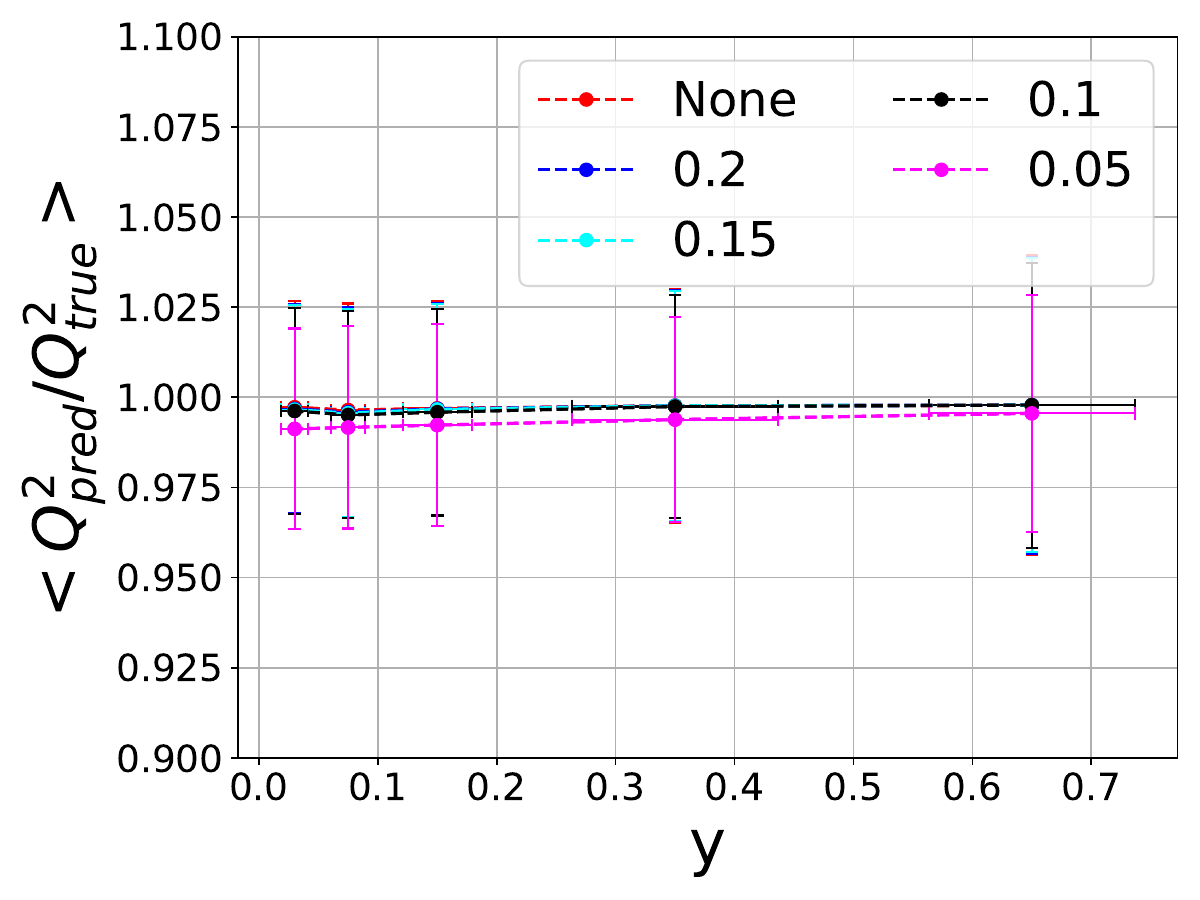} 
    \includegraphics[width=0.32\textwidth]{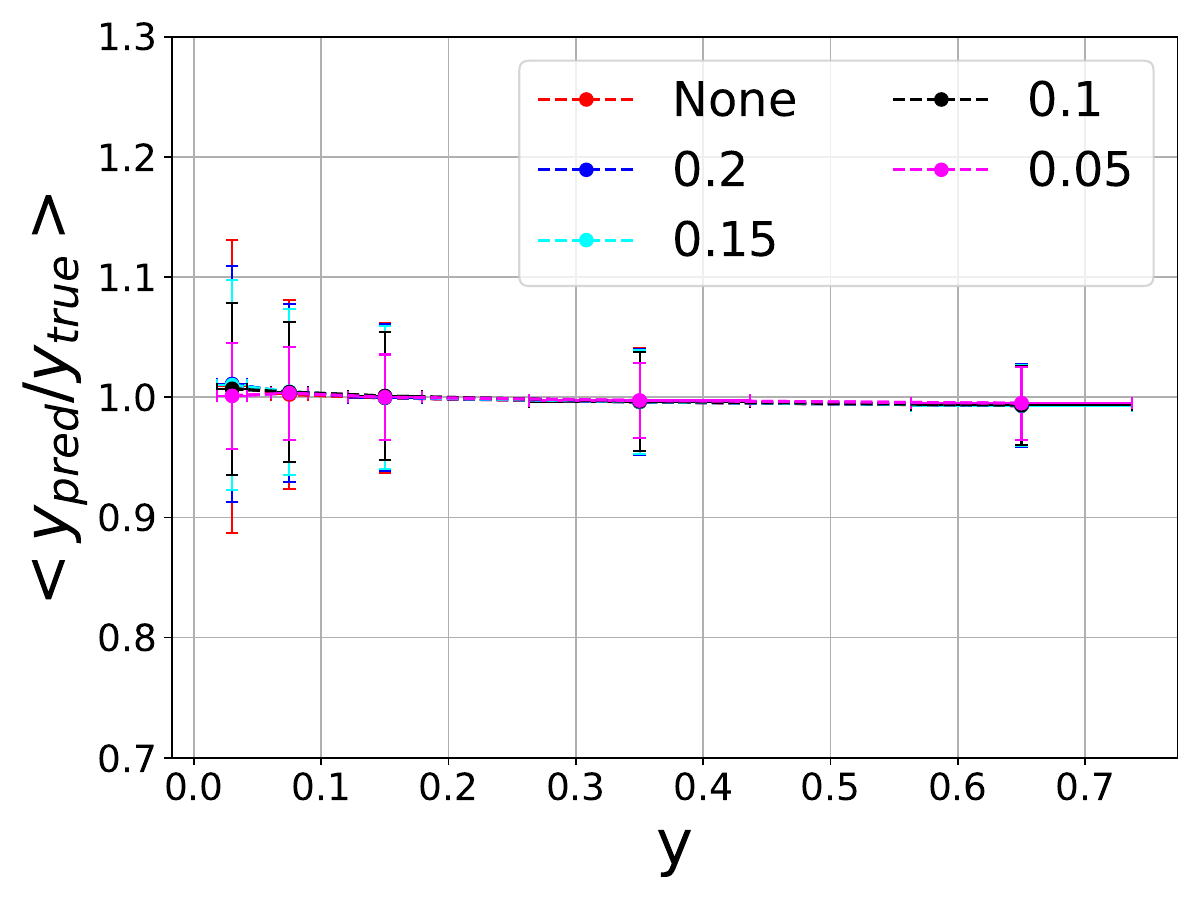} 
    \caption{Weighted average predicted-to-true ratio in bins of $y$ for ELUQ after applying a series of cuts at various thresholds on the relative weighted uncertainty of $x$, $Q^{2}$, $y$ to reject events.}
    \label{fig:weighted_unc_cuts}
\end{figure}

We also demonstrate how event-level UQ can be employed to assess the quality of events, retaining those with higher confidence and discarding events with more pronounced uncertainties.
Fig. \ref{fig:weighted_unc_cuts} shows the effect of cutting events with large relative uncertainty using different thresholds. 
By excluding events with higher uncertainty, we mitigate the observed drop in the predicted-to-true ratio for the variable \(x\).

\begin{figure}[!]
    \centering
    \includegraphics[width=0.45\textwidth]{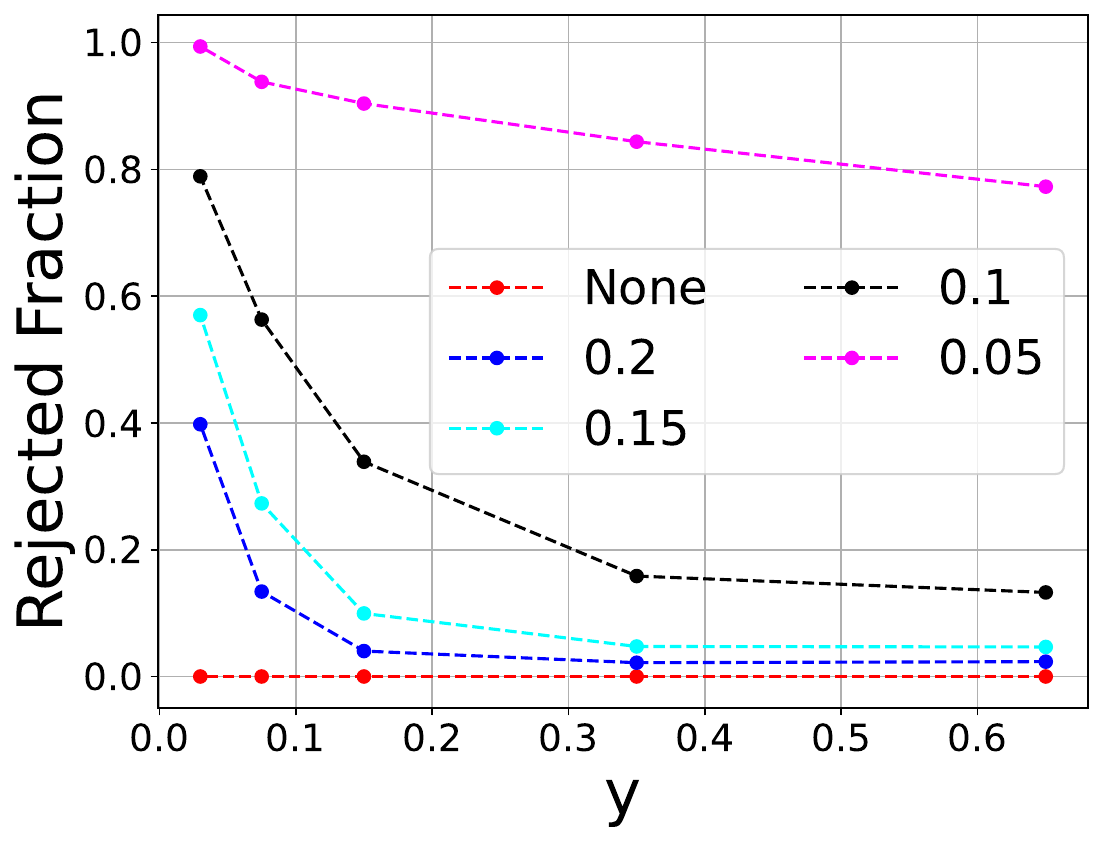} %
    \caption{Rejected Fraction in bins of $y$ after applying a series of boolean OR cuts at various thresholds on the relative uncertainty of $x$, $Q^{2}$, $y$. An event that does not pass any of these cuts is rejected.}
    \label{fig:stats-cascade}
\end{figure}

However, this approach results in a reduction of statistics. Figure \ref{fig:stats-cascade} illustrates the count of discarded events in relation to the severity of the cuts, segmented by bins of \(y\). The loosest cut removes 40\% of the events in the lowest bin in $y$, predominantly influenced by high aleatoric uncertainty in \(x\).

\section{Conclusions}\label{sec:Impacts} 

We present ELUQuant, a novel network that integrates Physics-Informed Bayesian Neural Networks with Flow-approximated Posteriors. This heralds a major leap in physics analyses, uniquely providing insights into both heteroskedastic aleatoric and epistemic uncertainties on an event-level basis. To our knowledge, this is a pioneering achievement in the field, realizing a long-sought benchmark.
 Validated by results from the H1 detector's DIS simulation at HERA, our work suggests promising future extensions to the upcoming EIC for extracting essential kinematic observables, which could be affected by radiation effects, and their associated uncertainties.
Closure tests support the reliability of the aleatoric and epistemic uncertainties extracted. For instance, aleatoric uncertainties align with the RMS of a DNN in y-regions where the regressed observables manifest as Gaussian distributions not affected by inaccuracy, centered at the expected mean from ground truth. Notably, epistemic uncertainty---originating from the same multivariate normal distribution characterizing the aleatoric term in the loss function---amplifies in response to increased inaccuracy with respect to ground truth.
 While the impact of ELUQ for DIS data is evident, its versatility extends to a broader range of event-level physics analyses. 
 The granularity ELUQ offers can revolutionize event filtering decision-making. Informed by uncertainties, it can mitigate true inaccuracies, showing promise in both data quality monitoring and anomaly detection.
In computational terms, our approach at inference showed an impressive rate of 10,000 samples/event within a mere 20 milliseconds on an RTX 3090, emphasizing real-world application viability.
In essence, ELUQuant's pioneering approach to event-level uncertainty quantification sets a new standard for comprehensive analyses in nuclear and particle physics.

%\clearpage

\section*{Acknowledgments}

% this is alphabetical reflecting author list
%
We thank the H1 Collaboration for allowing us to use the simulated MC event samples.

%\clearpage
\section*{References}
\bibliographystyle{iopart-num}
\bibliography{biblio}

%\clearpage
%\input{6_Appendix}

\end{document}